%% file: main.tex
\pdfoutput=1
\documentclass[10pt,twocolumn,letterpaper]{article}
\pdfinfoomitdate=1 \pdftrailerid{} \pdfsuppressptexinfo=-1
\usepackage[pagenumbers]{cvpr}

\input{preamble}
\usepackage{booktabs}
\usepackage{multirow}
\usepackage{amsmath,amssymb}
\usepackage{graphicx}
\usepackage{xcolor}
\usepackage{colortbl}
\definecolor{oursbg}{HTML}{F2F2F2}
\definecolor{deltagray}{HTML}{595959}
\definecolor{groupbg}{HTML}{E6E6E6}
\newcommand{\tgroup}[2]{\rowcolor{groupbg}\multicolumn{#1}{c}{\textbf{\textit{#2}}}}
\usepackage{pifont}
\usepackage{tabularx}
\definecolor{cvprblue}{rgb}{0.21,0.49,0.74}
\usepackage[pagebackref,breaklinks,colorlinks,allcolors=cvprblue]{hyperref}

\newcommand{\method}{GraphSeer}
\newcommand{\bench}{KRIR}
\newcommand{\cmark}{\ding{51}}
\newcommand{\xmark}{\ding{55}}
\newcommand{\cok}{\,\textcolor[HTML]{5B8A31}{\ding{51}}}
\newcommand{\cno}{\,\textcolor[HTML]{B95568}{\ding{55}}}
\newcommand{\cnote}[1]{{\color[HTML]{6B6A66}#1}}
\newcommand{\qa}[2]{\par\hangindent=0.6em\hangafter=1\textbf{#1:} #2}
\newcommand{\qcase}[4]{%
\begin{minipage}[t]{0.244\linewidth}\raggedright
{\footnotesize\textbf{Q:} #1}\par\vspace{-6pt}
\begin{minipage}[t]{0.31\linewidth}\vspace{0pt}#2\end{minipage}\hfill
\begin{minipage}[t]{0.66\linewidth}\vspace{-1pt}\raggedright\scriptsize\setlength{\baselineskip}{7.4pt}#3\end{minipage}%
\par\vspace{1.5pt}{\scriptsize\setlength{\baselineskip}{7.2pt}\cnote{#4}\par}
\end{minipage}}

\title{GraphSeer: Multimodal Knowledge Graph\\ Relational Image Retrieval with MLLMs}

\author{%
Linyu Li$^{1}$ \quad Zhi Jin$^{1,*}$ \quad Yuanpeng He$^{1}$ \quad Haoran Duan$^{2}$ \quad Dongming Jin$^{1}$\\
Wenchun Jing$^{1}$ \quad Xuan Zhang$^{1}$ \quad Gadeng Luosang$^{3}$ \quad Nyima Tashi$^{3}$\\[8pt]
\begin{minipage}{\dimexpr\textwidth-2\tabcolsep\relax}\centering\normalsize
$^{1}$School of Computer Science and Key Laboratory of High Confidence Software\\
Technologies, Peking University, Ministry of Education, Beijing, China\\[2pt]
$^{2}$School of Computer Science, Wuhan University, Wuhan, China\\[2pt]
$^{3}$School of Information Science and Technology and Tibetan Language Intelligence\\
National Key Laboratory, Tibet University, Lhasa, China\\[2pt]
$^{*}$Corresponding author\endgraf\vspace{-14pt}
\end{minipage}%
}

\begin{document}
\maketitle
\input{fig/teaser.tex}
\input{sec/0_abstract}
\input{sec/1_intro}
\input{sec/2_related}
\input{sec/3_task}
\input{sec/4_method}
\input{sec/5_experiments}
\input{sec/6_conclusion}
{
    \small
    \bibliographystyle{ieeenat_fullname}
    \bibliography{main}
}

\clearpage
\appendix
\renewcommand{\dbltopfraction}{0.9}
\setcounter{dbltopnumber}{3}
\widowpenalty=10000 \clubpenalty=10000
\raggedbottom
\input{sec/appendix}
\end{document}

%% file: preamble.tex
\usepackage{cuted} %

\usepackage{currfile} %

\usepackage{caption} %

%% file: fig/teaser.tex
\begin{strip}
\centering
\includegraphics[width=\linewidth]{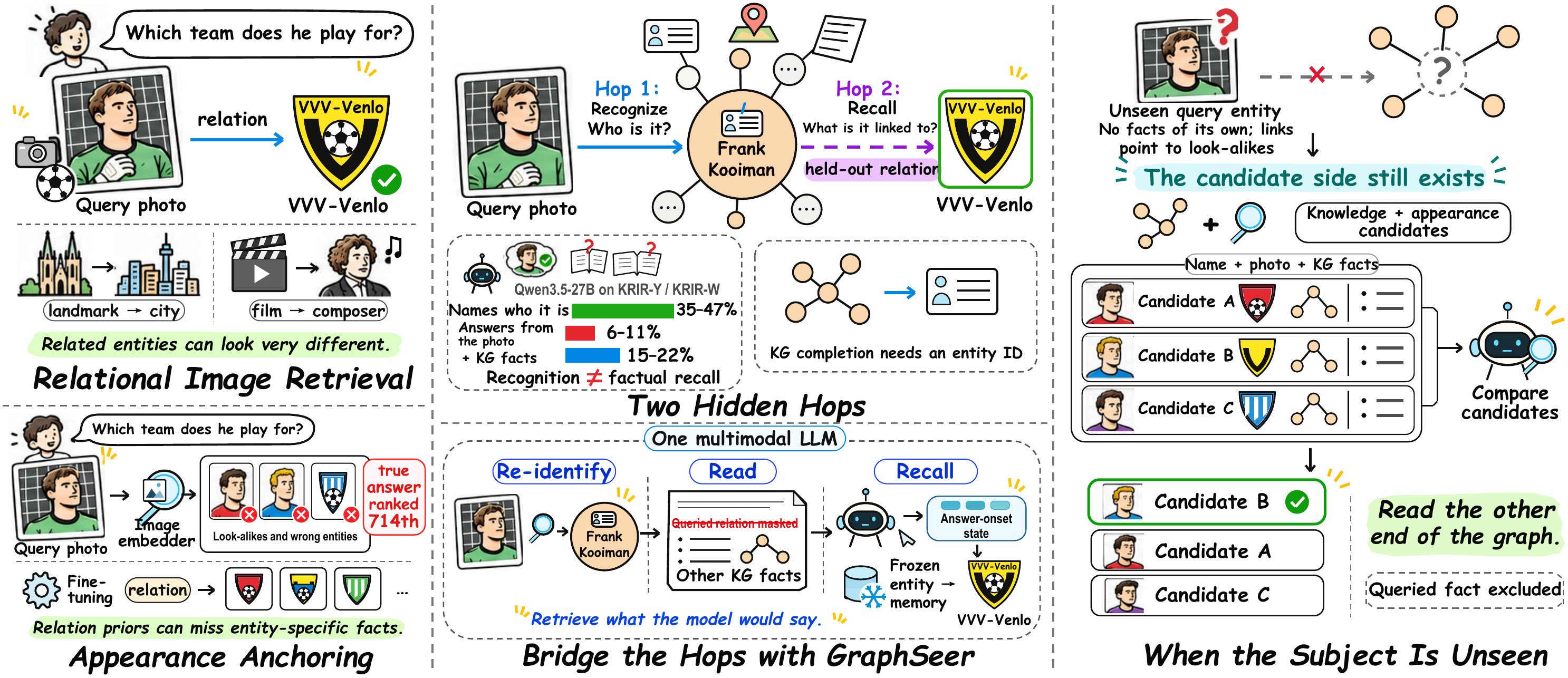}
\captionsetup{hypcap=false}
\captionof{figure}{\textbf{\bench{} and \method{}.} Given a photo and a relation, retrieve the MMKG entity linked to the depicted one, which rarely looks like it (top left). Image embedders return look-alikes, ranking the answer 714th here, and fine-tuning adds relation priors (bottom left). The task hides two hops, recognising who is shown and recalling what it is linked to (top middle). \method{} bridges them inside one MLLM (bottom middle) and, when the subject is unseen, verifies the candidates by their own facts (right). Drawn after a \bench{}-Y validation query.}
\captionsetup{hypcap=true}
\label{fig:teaser}
\end{strip}

%% file: sec/0_abstract.tex
\begin{abstract}
Multimodal knowledge graphs (MMKGs) store entities with photos and names and link them by facts.
MMKG methods predict links between known entity IDs, and image retrievers, from CLIP to MLLM embedders, return entities that look like a query photo or match its description.
Neither answers a common question about a new photo: which graph entity is related to the one shown, such as the team of a pictured player or the city of a landmark.
We call this \emph{knowledge graph relational image retrieval} (\bench{}): given a photo and a natural-language relation, retrieve the related MMKG entity, which rarely resembles the photo, whether the depicted entity is in the graph (known) or not (unseen).
We build three benchmarks from YAGO, Wikidata and in-the-wild OVEN photos.
The task hides two hops: recognising \emph{which} graph entity is shown, and recalling \emph{what} it is connected to.
\method{} bridges them inside one multimodal LLM: it links the photo to the graph, reads the facts of the best match, and retrieves with the hidden state at which the model is about to \emph{say} the answer.
An unseen entity has no facts of its own, but every candidate answer does, so a listwise verifier reads the other end of the graph: it re-ranks the top candidates by their facts and photos in one forward pass, and the facts matter far more than the photos.
With Qwen3.5-27B, \method{} improves MRR on known entities by 24.2--70.0\% over an MLLM embedder fine-tuned on the same data with the same budget.
When it also fuses and verifies the embedder's candidates, it improves MRR on unseen entities by 12.5--25.6\%, and on known entities it beats KG-completion models given the ground-truth entity ID.
\end{abstract}

%% file: sec/1_intro.tex
\section{Introduction}
\label{sec:intro}

\paragraph{Retrieval over multimodal knowledge graphs.}
Multimodal knowledge graphs (MMKGs) attach photos and names to entities and link them by facts~\cite{liu2019mmkg,xu2022mmrns}.
They are completed by predicting links between known entity IDs~\cite{xie2017ikrl,zhang2024native} and used to ground multimodal retrieval-augmented generation~\cite{park2026m3kgrag}, but are rarely searched from a new photo.
Yet a common question about a photo is relational: point a phone at a footballer and ask ``\emph{Which team does he play for?}''
The answer is a club, another entity of the graph connected to the player by a fact, and its crest shares nothing visual with the photo.
Other examples are the director of a film from its poster, the city of a landmark and the label of an album cover, yet the benchmarks of current image retrievers do not test such questions.
We formalise \textbf{knowledge graph relational image retrieval} (\bench{}) over MMKGs: given a photo and a natural-language relation, rank the entities of an MMKG, each with a photo and a name, so that the entities linked to the depicted one by that relation come first (\cref{fig:teaser}).
The depicted entity may be in the graph with the queried fact missing (transductive), or absent from it altogether (inductive).

\paragraph{Why existing methods fail.}
Take the footballer Frank Kooiman in \cref{fig:teaser}.
Qwen3-VL-Embedding-8B~\cite{li2026qwen3vlemb}, a top model on MMEB-V2, ranks his club VVV-Venlo 714th.
Composed image retrieval~\cite{liu2021cirr,saito2023pic2word,zhang2024magiclens} and MLLM embedders~\cite{jiang2025vlm2vec,zhang2025gme} assume that the query \emph{describes} the target, which is not true of a relational target.
Reaching it takes two hidden hops over the MMKG: recognising \emph{which} graph entity is in the photo, and recalling \emph{what} that entity is connected to.
MMKG completion handles the second hop, but it needs the entity ID and uses images only as features of known entities~\cite{xie2017ikrl,zhang2024native,li2025tsam}.
Multimodal LLMs attempt both, but only generatively and unreliably~\cite{ashok2025vlmrecall,venhoff2025toolate}: Qwen3.5-27B names the depicted entity for 35--47\% of our queries, yet recalls a held-out answer from the photo for only 6--11\% (\cref{sec:diag}).
Fine-tuning an embedder on the graph helps, but the embedder learns which answers a relation tends to have rather than the facts of the depicted entity.

\paragraph{Our approach.}
The MMKG itself provides what both hops need: its photos let one MLLM re-identify the depicted entity, its facts say what that entity is connected to, and its photos and names form the gallery.
\textbf{\method{}} (\cref{fig:method}) \emph{re-identifies} Kooiman among the MMKG entities, \emph{reads} the verbalised facts of the match (facts of the queried relation are masked) and \emph{recalls}: the query embedding is the hidden state at the \emph{answer onset}, where the model would start the answer's name, matched against a frozen entity memory built by the same MLLM.
Trained on the linker's real links, a third of which are wrong, it learns when to trust the graph.
It then re-scores the top candidates, its own or an appearance retriever's, by the likelihood of their names.
An unseen entity has no facts of its own and its links point to look-alikes, but the answer is still an MMKG entity with its own name, photo and facts, so \method{}'s listwise verifier puts the top-20 candidates with these in one prompt and reads a distribution over them from a single token.

\paragraph{Contributions.}
\textbf{(1)} We introduce \bench{}, relational image retrieval over MMKGs, with three benchmarks built on the YAGO and Wikidata MMKGs of MMRNS~\cite{xu2022mmrns} and on Wikidata5M with in-the-wild OVEN photos~\cite{hu2023oven,wang2021kepler}, and show that current retrievers anchor on appearance while fine-tuning teaches them relation priors rather than entity facts.
\textbf{(2)} We propose \method{}, which uses the MMKG at both ends of a query: it retrieves with answer-onset states of an MLLM that reads the linked entity's facts, and its listwise verifier reads the candidates' own names, photos and facts.
\textbf{(3)} With Qwen3.5-27B, \method{} improves MRR over a fine-tuned Qwen3-VL-Embedding by 24.2--70.0\% on known entities of all three benchmarks and, on \bench{}-W/OVEN, beats KG completion given the ground-truth entity ID; fusing both retrievers and verifying their top candidates gives 39.1/46.7/56.1 MRR on known and 16.2/30.6/38.0 on unseen entities, 12.5--25.6\% above the fine-tuned embedder on unseen ones.

%% file: sec/2_related.tex
\section{Related Work}
\label{sec:related}

\paragraph{Multimodal knowledge graphs.}
MMKGs attach images and names to the entities of a KG~\cite{liu2019mmkg}.
MMKG completion predicts links between entity IDs with images as side features~\cite{xie2017ikrl,xu2022mmrns,zhang2024native,zhang2025mygo,li2025tsam}, also inductively or with LLMs~\cite{zhao2024cmr,efthymiou2026vlkge,zhang2024kopa,wang2022simkgc}, and MMKGs ground multi-hop retrieval-augmented generation~\cite{park2026m3kgrag}, but none of these takes a new photo as the query.
The closest task, ImageGraph~\cite{onoro2019imagegraph}, answers $(\text{image}, r, ?)$ queries, including images of new entities, with VGG features and fixed relation IDs; \bench{} states the relation in natural language and ranks entities by photo and name.

\paragraph{Composed and instruction-based retrieval.}
Composed image retrieval finds a target image from a reference image and a modification text~\cite{wu2021fashioniq,liu2021cirr,vaze2023genecis}, zero-shot through pseudo-words, external knowledge or LLM reasoning~\cite{saito2023pic2word,baldrati2023searle,suo2024keds,tang2025osrcir} or with MLLMs~\cite{yang2026recall}, and MagicLens and MegaPairs train on open-ended instructions~\cite{zhang2024magiclens,zhou2025megapairs}.
Their targets stay close to the query in appearance; a \bench{} target is a different entity linked by a fact.

\paragraph{MLLM embedders and rerankers.}
MLLM embedders come from prompting~\cite{jiang2024e5v} or contrastive tuning~\cite{wei2024uniir,jiang2025vlm2vec,zhang2025gme,lin2025mmembed,liu2025lamra,li2026qwen3vlemb,liu2026rematch}, and break down when retrieval needs reasoning~\cite{su2025bright,zhou2025mr2bench,li2026rmir}; KBMR~\cite{xu2026kbmr} aligns a photo with its own entity rather than with look-alikes.
\method{} instead embeds the state at which the MLLM is about to say the answer.
LLM rerankers order a candidate list in one pass~\cite{sun2023rankgpt,pradeep2023rankzephyr,reddy2024first,liu2025lamra} by matching candidate content to the query, whereas our verifier reads graph evidence on the candidate's side.

\paragraph{Visual entity recognition and MLLM recall.}
OVEN, InfoSeek and Encyclopedic-VQA link an image to its own entity or ask about one of its attributes~\cite{hu2023oven,chen2023infoseek,mensink2023evqa,caron2024ger,xiao2024autover,yan2024echosight,ning2026wikiclip}, and retrieved entity names aid open-world captioning~\cite{li2024evcap}.
Knowledge-based VQA retrieves entities or documents to answer a question about the photo~\cite{chen2024vtqa,cocchi2025reflectiva,yang2025qacap,compagnoni2026reag}, but generates a text answer rather than ranking KG entities.
VLMs recall facts far less reliably about a pictured than a named subject~\cite{cohen2025popvqa,ashok2025vlmrecall}, because the visual entity forms too late for the LLM's recall circuitry~\cite{venhoff2025toolate,geva2023dissecting,hernandez2024linearity}.
\method{} therefore links the photo into the graph, so that the subject enters the model as text.

%% file: sec/3_task.tex
\section{Relational Image Retrieval over Multimodal Knowledge Graphs}
\label{sec:task}

\subsection{Task}
Let $\mathcal{G}=(\mathcal{E},\mathcal{R},\mathcal{T})$ be a multimodal knowledge graph (MMKG) in which every entity $e$ has a name $n_e$ and a canonical photo $I_e$.
A query is a pair $(I_q, r)$: a photo of some entity $h$, whose identity is \emph{not} given to the system, and a relation $r$ expressed in natural language, \eg ``\emph{Which team does the entity shown in the image play for?}''.
The system ranks the gallery $\{(I_e, n_e)\}_{e\in\mathcal{E}}$, and every $t$ with $(h,r,t)$ true counts as a correct answer.
Inverse relations are separate queries (``\emph{Who plays for \ldots?}''), so 1-to-N and N-to-1 relations both appear.
We report filtered MRR and Hits@1/3/10 over the full gallery: other true answers of the same query are removed from the ranking, and ties count half.

In the \textbf{transductive} regime, $h$ is a KG entity, and the queried fact is missing from the graph.
In the \textbf{inductive} regime, $h$ is an \emph{unseen} entity: all of its triples, its name and its gallery entry are removed, and only its photo arrives at test time.
We also call the two regimes \emph{known} and \emph{unseen} entities.
KG-embedding methods have no trained embedding for an unseen entity.

\begin{figure*}[t]
\centering
\includegraphics[width=\linewidth]{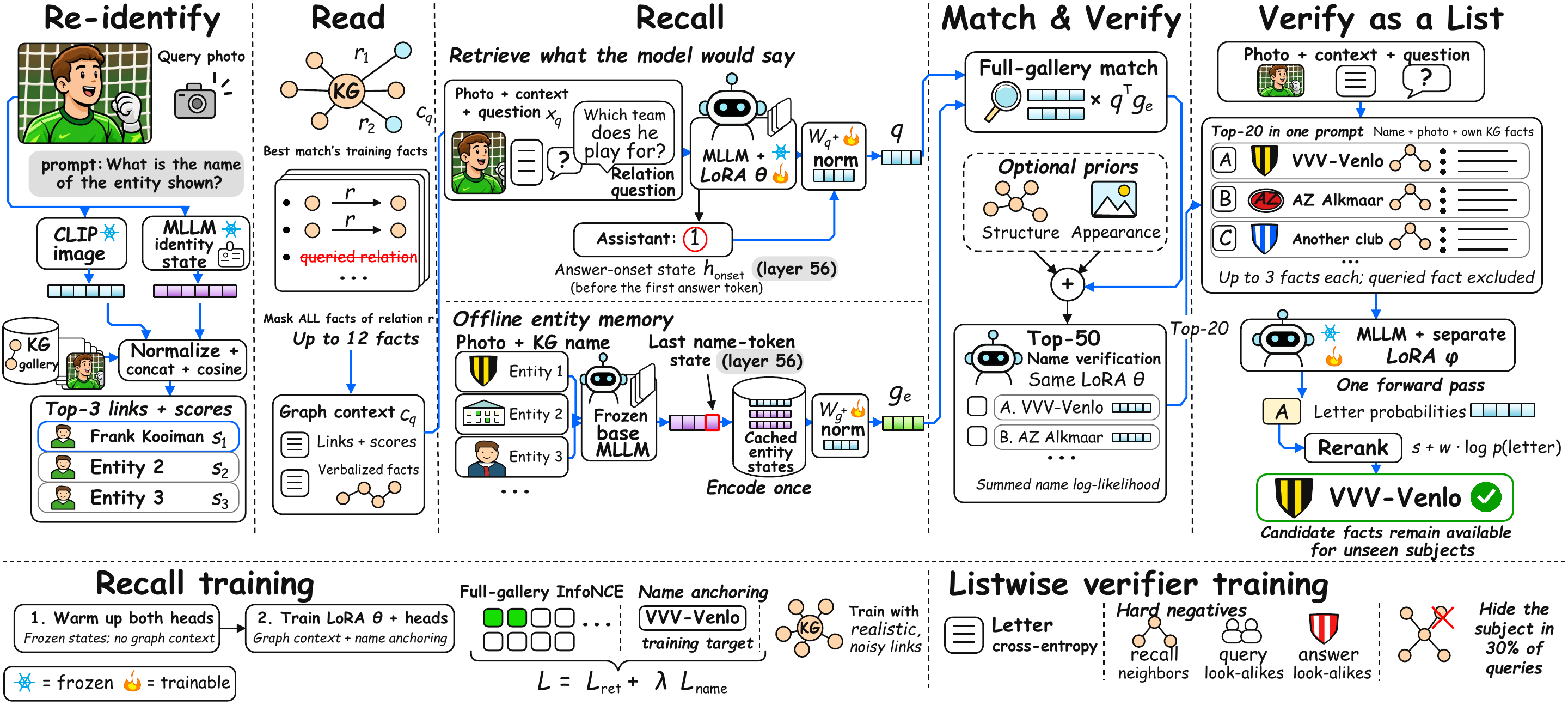}
\caption{\textbf{\method{} overview.} \emph{Re-identify}: frozen CLIP and MLLM identity states link the photo to its top-3 KG entities. \emph{Read}: the best match's facts (none of the queried relation) and the links form the graph context $c_q$. \emph{Recall}: the LoRA-tuned MLLM reads photo, context and question, and its answer-onset state becomes the query $q$; a frozen memory encodes every entity once ($g_e$). \emph{Match \& Verify}: full-gallery matching, optionally fused with structural and appearance priors, then name verification of the top-50 with the same LoRA. \emph{Verify as a List}: a separate LoRA re-ranks the top-20 from each candidate's name, photo and own facts (\cref{sec:listwise}). Bottom: training of recall and of the listwise verifier.}
\label{fig:method}
\end{figure*}

\subsection{Benchmarks}
We build \bench{}-Y and \bench{}-W on the MMKG-completion benchmarks MKG-Y (YAGO, 28 relations) and MKG-W (Wikidata, 169 relations) of MMRNS~\cite{xu2022mmrns}, whose entities come with up to three web images.
An entity's first image is its gallery photo, and its query photo is another image that is not a near-duplicate, so re-identification cannot rely on exact matching.
For the inductive regime, 10\% of the entities that head an evaluation fact are removed from the graph and the gallery; all facts of a removed entity, in any split, become inductive queries, and the remaining triples form one training graph for both regimes.

\paragraph{\bench{}-OVEN: photos taken in the wild.}
On \bench{}-Y/W, query and gallery photos come from the same web crawl.
To test query photos from another source, we build a third benchmark from the entity split of OVEN~\cite{hu2023oven}, whose images come from recognition datasets (landmarks, species, aircraft, food, \etc) and are labelled with the depicted Wikidata entity.
Facts from Wikidata5M~\cite{wang2021kepler} and the entities' Wikipedia photos~\cite{wei2024uniir} form the MMKG, and 30\% of its eligible entities are unseen.
Query photos differ from gallery photos in viewpoint, scale and clutter: top-1 re-identification drops from 61--71\% on \bench{}-Y/W to 38\% on \bench{}-OVEN.
\Cref{tab:stats} summarises the three benchmarks.

\begin{table}[t]
\centering
\small
\setlength{\tabcolsep}{3pt}
\resizebox{\linewidth}{!}{%
\begin{tabular}{lrrr}
\toprule
 & \bench{}-Y & \bench{}-W & \bench{}-OVEN \\
\midrule
Source KG / relations & YAGO / 28 & Wikidata / 169 & Wikidata5M / 49 \\
Query photos & held-out web & held-out web & in the wild \\
Gallery entities (photo + name) & 13{,}782 & 8{,}758 & 7{,}644 \\
Training triples & 19{,}238 & 31{,}381 & 13{,}624 \\
Unseen entities (val / test) & 232 / 466 & 173 / 347 & 142 / 284 \\
Transductive queries (val / test) & 3{,}618 / 3{,}610 & 2{,}014 / 2{,}060 & 597 / 615 \\
Inductive queries (val / test) & 400 / 807 & 361 / 702 & 331 / 659 \\
Test facts (trans. / ind.) & 4{,}046 / 1{,}742 & 2{,}523 / 1{,}423 & 677 / 1{,}250 \\
\bottomrule
\end{tabular}}
\caption{\bench{} statistics. A query is a (photo, relation) pair and may have several correct answers (facts).}
\label{tab:stats}
\end{table}

\subsection{Where the two hops break}
\label{sec:diag}
The top middle of \cref{fig:teaser} measures the problem with Qwen3.5-27B~\cite{qwen35} on 500 validation queries per benchmark, counting an answer only if it matches a held-out answer.
The model names the depicted entity for 35\%/47\% of the queries (\bench{}-Y/W) but answers from the photo for only 6\%/11\%, below the 8\%/14\% reached when the question names the entity, the two-hop gap of~\cite{venhoff2025toolate}.
The verbalised facts of the visually linked entity raise this to 15\%/22\%.
A KG embedding, in contrast, needs the entity ID: ComplEx reaches 36.2/37.5 MRR on known entities with the oracle ID but is near zero on unseen ones.
Once the photo is linked, the graph supplies what recall from pixels misses, so \method{} links first and reads second.

%% file: sec/4_method.tex
\section{\method{}: Re-identify, Read,\texorpdfstring{\protect\\}{ } Recall, Verify}
\label{sec:method}

\Cref{fig:method} shows \method{} on the running example.
Each step draws on a different part of the MMKG.
\method{} links the photo to candidate entities through their photos (\emph{Re-identify}), writes the facts of the best match into the prompt (\emph{Read}), and embeds the query as the MLLM state at the answer onset, which is matched against a frozen memory built from the entities' photos and names (\emph{Recall}).
The top candidates are then verified by the likelihood of their names under the same model and by a listwise verifier that reads each candidate's own facts and photo (\emph{Verify}, \cref{sec:listwise}).

\subsection{Answer-onset embeddings}
Let $f_\theta$ be an MLLM and $x$ a multimodal prompt ending with the assistant turn.
Common MLLM embedders take the final state of a \emph{describe} prompt (``summarise the image in one word'')~\cite{jiang2024e5v} or of an instruction-wrapped input~\cite{jiang2025vlm2vec,zhang2025gme}.
For relational retrieval this is the wrong state, because it encodes what the query \emph{looks like}.
Instead, we ask the relational question with the instruction to answer by name only, and read a late-layer state $\mathbf{h}^{\ell}$ at the \emph{answer onset}, the position whose next token begins the answer:
\begin{equation}
  \mathbf{q}=\mathrm{norm}\big(W_q\,\mathbf{h}^{\ell}_{\text{onset}}(x_q)\big),\qquad
  x_q=[\,I_q;\ c_q;\ \text{question}(r)\,].
\end{equation}
Here $c_q$ is the graph context (\cref{sec:read}).
At this position the model has to commit to the answer it is about to produce, so its late-layer states encode the identity and type of that answer, and retrieval follows what the model would say.
Even without training, answer-onset matching beats describe-style prompting with the same backbone from layer 40 on (\cref{fig:analysis}(a)).

\subsection{Re-identify and Read: graph context}
\label{sec:read}
\paragraph{Re-identify.}
For each query photo, we score every gallery entity by the cosine similarity of concatenated, normalised CLIP image embeddings~\cite{radford2021clip} and mean-centred MLLM identity states (the answer-onset state of ``\emph{What is the name of the entity shown?}''), which add recognition, including visible text, to instance appearance.
We keep the top-$3$ candidates $\{(e_k,s_k)\}$.
\paragraph{Read.}
The context $c_q$ lists the candidates with their scores and up to 12 training facts of the best candidate $e_1$, verbalised with relation templates (\eg ``\emph{Where was it born? Leiden}'').
Facts of the queried relation $r$ are dropped, so the model cannot copy a known answer.
Linking turns the subject into \emph{text}, the route through which LLM recall works best~\cite{ashok2025vlmrecall,venhoff2025toolate}, and adds structural knowledge the backbone never memorised.

\subsection{Recall and Match: a frozen entity memory}
Every gallery entity is encoded \emph{once} by the frozen MLLM.
The prompt contains the entity's photo, the identity question, and its KG name given as the answer, and we read a late-layer state at the last name token:
$\mathbf{g}_e=\mathrm{norm}\big(W_g\,\mathbf{h}^{\ell}_{\text{name}}(I_e, n_e)\big)$.
This memory holds the MLLM's own representation of each entity, enriched by the name even when the photo is not recognised.
Because it is frozen, only LoRA~\cite{hu2022lora} on the query side and the linear heads $W_q,W_g$ are trained.
The contrastive loss is computed against the \emph{entire} gallery at every step, with no hard-negative mining or gallery re-encoding.

\subsection{Training}
Training queries are $(h,r)$ groups of the training graph with answers $\mathcal{A}_q$, posed with the held-out photo of $h$.
\paragraph{Full-gallery multi-positive InfoNCE.}
With $z_{qe}=\tau\,\mathbf{q}^{\top}\mathbf{g}_e$,
\begin{equation}
\mathcal{L}_{\text{ret}}=\frac{1}{|\mathcal{A}_q|}\sum_{a\in\mathcal{A}_q}-\log\frac{\exp(z_{qa})}{\exp(z_{qa})+\sum_{e\notin\mathcal{A}_q}\exp(z_{qe})},
\end{equation}
so each training answer competes only with non-answers; validation and test answers are never used.
\paragraph{Name anchoring.}
Contrastive tuning tends to pull an embedding toward whatever separates the gallery most easily, here appearance.
We keep the answer-onset state tied to verbalised knowledge by teacher-forcing the KG name of one sampled answer $a$ after the onset:
\begin{equation}
\mathcal{L}_{\text{name}}=-\sum_{j}\log p_\theta\big(n_a^{(j)}\mid x_q,\ n_a^{(<j)}\big).
\end{equation}
Names serve only as training targets; at test time the query names only the linked candidates.
\paragraph{Learning from realistic links.}
An unseen entity is always linked to a \emph{wrong} KG entity at test time, so we never train with oracle links but with the same linker run on the training photos, whose top-1 is wrong for 39\%\,/\,29\% of training queries on \bench{}-Y\,/\,\bench{}-W.
Seeing misleading contexts with their match scores, the model learns when to trust the graph, with no hand-set gate (\cref{tab:abl}(a)).
The total loss is $\mathcal{L}=\mathcal{L}_{\text{ret}}+\lambda\,\mathcal{L}_{\text{name}}$ with $\lambda=1$, chosen on validation from $\{0.5,1\}$.

Since randomly initialised heads would need far more updates than a LoRA run affords, stage 1 fits $W_q,W_g$ with $\mathcal{L}_{\text{ret}}$ on frozen answer-onset states (minutes), and stage 2 trains LoRA and the heads jointly.

\subsection{Verification}
\label{sec:listwise}
\input{tables/main_compact}
\paragraph{Name likelihood.}
An embedding compresses the anticipated answer into one vector, but the fine-tuned model can also \emph{say} it, and name anchoring trains exactly $p_\theta(n_a\mid x_q)$.
At test time we re-score each of the top-$K$ ($K{=}50$) retrieved entities $t$ by the log-likelihood of its full KG name after the answer onset,
\begin{equation}
s(t)=s_{\text{rec}}(t)+\eta\textstyle\sum_j\log p_\theta\big(n_t^{(j)}\mid x_q,n_t^{(<j)}\big),
\end{equation}
where $s_{\text{rec}}(t)=z_{qt}$, and $\eta$, $K$ and the aggregation (sum or mean over tokens) are chosen on validation; this costs $K$ short forward passes with the same weights.

\paragraph{Priors.}
Other retrievers can be added at inference without retraining \method{}.
With $p(e_k\mid I_q)\propto\exp(s_k/T)$ over the linked candidates, we add $\beta\,s_{\text{kg}}(t)$ to the log-softmax of the recall scores, where the \emph{structural prior} $s_{\text{kg}}(t)=\log\sum_k p(e_k\mid I_q)\,\mathrm{softmax}_t\,\psi(e_k,r,t)$ uses ComplEx scores $\psi$.
An \emph{appearance prior} $\gamma\log\mathrm{softmax}_t\,s_{\text{app}}(I_q,r,t)$ from an embedder, zero-shot or fine-tuned on \bench{}, is added in the same way.
$T$, $\beta$ and $\gamma$ are selected on validation, and the top-$K$ of the fused ranking is verified, whichever retriever proposed each candidate.

\paragraph{Listwise verification.}
Recall and name verification read the facts of the re-identified entity, which for an unseen entity are missing or belong to a look-alike.
Every candidate answer, however, has its own facts and photo, which tell what kind of entity it is and what it is connected to, and a list lets the model compare the candidates.

The verifier shows the MLLM the top-$M$ candidates of a ranking at once ($M{=}20$).
After the query photo, the graph context and the relation question, candidate $i$ is written as a letter $y_i$, its KG name, its photo (64 visual tokens) and up to three of its training facts, those that mention a re-identified entity first.
One forward pass gives $p_\phi(y_i\mid x)$ over the $M$ letters at the answer onset, read from a single token as in FIRST~\cite{reddy2024first}, and
\begin{equation}
s'(t_i)=s(t_i)+w\log p_\phi(y_i\mid x),\qquad i\le M,
\end{equation}
with $w$ chosen on validation.
The verifier is its own LoRA $\phi$ on the frozen backbone, trained with cross-entropy over the letters on training triples.
Each list holds one answer and $M{-}1$ negatives from three sources: stage-1 recall neighbours (entities the relation makes plausible), look-alikes of the query photo (the appearance shortcut) and look-alikes of the answer, in random order.
Candidate facts never include the queried fact.
With probability $0.3$ the depicted entity is hidden: it is removed from the links, together with every candidate fact that mentions it, so that the training query looks like an unseen one.

%% file: tables/main_compact.tex
\begin{table*}[t]
\centering
\renewcommand{\arraystretch}{0.9}
\small
\setlength{\tabcolsep}{4.2pt}
\resizebox{0.9\textwidth}{!}{%
\begin{tabular}{l cccc cccc cccc}
\toprule
 & \multicolumn{4}{c}{\bench{}-Y (YAGO)} & \multicolumn{4}{c}{\bench{}-W (Wikidata)} & \multicolumn{4}{c}{\bench{}-OVEN (in the wild)} \\
\cmidrule(lr){2-5}\cmidrule(lr){6-9}\cmidrule(lr){10-13}
 & \multicolumn{2}{c}{Known} & \multicolumn{2}{c}{Unseen} & \multicolumn{2}{c}{Known} & \multicolumn{2}{c}{Unseen} & \multicolumn{2}{c}{Known} & \multicolumn{2}{c}{Unseen} \\
\cmidrule(lr){2-3}\cmidrule(lr){4-5}\cmidrule(lr){6-7}\cmidrule(lr){8-9}\cmidrule(lr){10-11}\cmidrule(lr){12-13}
Method & MRR & H@1 & MRR & H@1 & MRR & H@1 & MRR & H@1 & MRR & H@1 & MRR & H@1 \\
\midrule
\tgroup{13}{Zero-shot embedders} \\
CLIP-L/14~\cite{radford2021clip} (image + relation text) & 1.0 & 0.2 & 1.0 & 0.6 & 2.4 & 0.6 & 2.9 & 1.6 & 4.1 & 1.9 & 4.4 & 2.3 \\
SigLIP2-so400m~\cite{tschannen2025siglip2} (image + text) & 0.7 & 0.1 & 0.6 & 0.2 & 2.3 & 0.9 & 2.3 & 1.1 & 4.4 & 1.9 & 4.5 & 2.4 \\
BGE-VL-L (MegaPairs)~\cite{zhou2025megapairs} & 0.3 & 0.0 & 0.4 & 0.1 & 0.9 & 0.3 & 0.8 & 0.3 & 1.8 & 0.4 & 2.9 & 1.4 \\
GME-Qwen2-VL-7B~\cite{zhang2025gme} & 5.7 & 1.4 & 6.9 & 4.0 & 12.2 & 4.5 & 13.4 & 8.9 & 24.5 & 16.8 & 19.4 & 14.2 \\
Qwen3-VL-Embedding-8B~\cite{li2026qwen3vlemb} & 8.2 & 4.0 & 10.3 & 7.3 & 14.0 & 6.5 & 15.7 & 11.3 & 23.5 & 14.6 & 20.3 & 13.2 \\
\midrule
\tgroup{13}{Trained on the \bench{} graph (same triples, full-gallery InfoNCE)} \\
CLIP4Cir Combiner~\cite{baldrati2022clip4cir} (CLIP-L/14) & 15.3 & 9.9 & 7.8 & 4.4 & 28.0 & 21.1 & 20.4 & 14.8 & 34.6 & 25.6 & 25.2 & 17.4 \\
Qwen3-VL-Embedding-8B, LoRA fine-tuned & 21.3 & 15.3 & 12.9 & 8.6 & 35.5 & 28.1 & 27.2 & 20.7 & 42.3 & 31.9 & 32.5 & 23.4 \\
Fine-tuned emb. + name verifier (untuned Qwen3.5-27B) & 25.9 & 20.4 & 14.3 & 10.3 & 38.3 & 31.5 & 29.2 & 23.1 & 42.6 & 31.9 & 33.0 & 24.0 \\
Fine-tuned emb. + name verifier (\method{}) & 28.1 & 23.3 & 13.5 & 9.5 & 39.9 & 33.7 & 28.9 & 23.4 & 43.3 & 33.1 & 32.4 & 23.4 \\
Fine-tuned emb. + listwise verifier (\method{}) & 27.4 & 22.3 & \underline{15.7} & \underline{11.7} & 41.4 & 35.6 & \textbf{30.9} & \textbf{25.0} & 50.6 & 41.2 & \underline{37.2} & \underline{27.8} \\
\midrule
\tgroup{13}{Generative MLLM (Qwen3.5-27B) + name matching} \\
Generate answer $\to$ match & 6.8 & 5.4 & 5.4 & 3.9 & 14.9 & 12.3 & 12.1 & 10.0 & 20.2 & 13.9 & 17.4 & 13.0 \\
Generate with graph context (our linker) & 16.7 & 15.1 & 6.0 & 4.5 & 23.5 & 20.3 & 12.0 & 9.7 & 27.1 & 19.9 & 16.5 & 12.3 \\
\midrule
\tgroup{13}{Knowledge-graph embedding} \\
ComplEx-N3~\cite{lacroix2018n3}, re-ID link & 23.4 & 20.1 & 5.9 & 4.0 & 32.1 & 25.8 & 17.4 & 13.2 & 25.3 & 18.9 & 16.3 & 12.1 \\
\textcolor{gray}{ComplEx-N3, oracle entity ID$^\dagger$} & 36.2 & 32.8 & 0.0 & 0.0 & 37.5 & 31.0 & 1.1 & 0.0 & 25.6 & 19.1 & 3.4 & 1.8 \\
\textcolor{gray}{NativE~\cite{zhang2024native}, oracle entity ID$^\dagger$} & 36.9 & 32.9 & 4.0 & 1.8 & 37.8 & 30.0 & 13.5 & 9.4 & 34.5 & 26.4 & 17.4 & 11.7 \\
\midrule
\tgroup{13}{\method{} (ours, Qwen3.5-27B)} \\
\rowcolor{oursbg}Stage 1: frozen answer-onset states & 21.4 & 15.5 & 9.0 & 5.5 & 31.9 & 24.8 & 20.5 & 15.8 & 34.8 & 25.4 & 19.6 & 12.8 \\
\rowcolor{oursbg}Recall only & 31.3 & 26.0 & 9.8 & 6.1 & 37.7 & 30.8 & 21.9 & 17.0 & 44.1 & 35.5 & 20.7 & 14.1 \\
\rowcolor{oursbg}Recall + name verification & 34.4 & 29.5 & 11.9 & 8.2 & 40.8 & 34.0 & 24.2 & 18.7 & 45.0 & 36.8 & 22.9 & 15.9 \\
\rowcolor{oursbg}Recall + name + listwise verification & 36.2 & \underline{31.4} & 14.2 & 10.8 & \underline{44.1} & \underline{37.9} & 27.1 & 21.9 & \underline{53.7} & \underline{45.6} & 29.2 & 22.7 \\
\rowcolor{oursbg}Recall + priors + name verification & \underline{36.9} & 31.2 & 14.5 & 10.4 & 43.3 & 36.1 & 28.8 & 22.1 & 46.8 & 35.6 & 31.0 & 21.8 \\
\rowcolor{oursbg}Recall + priors + name + listwise (full system) & \textbf{39.1} & \textbf{33.9} & \textbf{16.2} & \textbf{12.3} & \textbf{46.7} & \textbf{40.3} & \underline{30.6} & \underline{24.0} & \textbf{56.1} & \textbf{46.1} & \textbf{38.0} & \textbf{28.6} \\
\bottomrule
\end{tabular}}
\caption{\textbf{Main results} on the \bench{} test sets (filtered MRR and Hits@1 over the full gallery, \%). Known/unseen: transductive/inductive queries. Verifiers in the fine-tuned embedder rows re-rank its top-20. $^\dagger$Uses the ground-truth entity ID of the query photo (untrained for unseen entities). H@3 and H@10 are in the appendix. Best in bold, second best underlined.}
\label{tab:main}
\end{table*}

%% file: sec/5_experiments.tex
\section{Experiments}
\label{sec:exp}

\subsection{Setup}
\paragraph{Implementation.}
\method{} uses Qwen3.5-27B~\cite{qwen35} with thinking disabled, so the answer onset is the first position after the empty think block.
LoRA (rank 16) is applied to the language model and the vision tower stays frozen.
Stage 2 fine-tunes for 4/6/10 epochs on \bench{}-Y/W/OVEN with a global batch of 64 on 8 A800 GPUs; the entity memory and the query both use layer 56.
The listwise verifier (rank-16 LoRA, $M{=}20$, a $256^2$ photo and 3 facts per candidate) is trained for 1/1/2 epochs on \bench{}-Y/W/OVEN, fixed in advance.
Checkpoints, fusion weights and layers are selected on validation only (details in the appendix), and differences are relative changes of MRR.

\paragraph{Baselines.}
We evaluate dual encoders (CLIP-L/14, SigLIP2) with the composed query $\mathrm{norm}(\mathbf{v}_{\text{img}}+\mathbf{v}_{\text{text}})$ of zero-shot composed image retrieval (CIR)~\cite{saito2023pic2word}, the CIR model BGE-VL~\cite{zhou2025megapairs}, and the MLLM embedders GME-Qwen2-VL-7B~\cite{zhang2025gme} and Qwen3-VL-Embedding-8B~\cite{li2026qwen3vlemb}.
We also train the CLIP4Cir Combiner~\cite{baldrati2022clip4cir} and Qwen3-VL-Embedding-8B (LoRA) on exactly our training triples with our full-gallery InfoNCE, tuned on validation with the same tuning budget as ours.
The fine-tuned embedder is thus our recipe with its own pooled query state in place of the answer-onset state of a graph-reading MLLM.
We further report generative answering with Qwen3.5-27B matched to gallery names, ComplEx-N3~\cite{trouillon2016complex,lacroix2018n3} with our re-identification (re-ID) links or the oracle entity ID, the MMKG-completion model NativE~\cite{zhang2024native} with the oracle ID, and our method without fine-tuning.

\subsection{Main results}
\label{sec:main}
\input{fig/main_analysis.tex}
\input{tables/ablation_merged}
\input{fig/qual_cases}

\paragraph{Appearance, generation and entity IDs fall short.}
\Cref{tab:main} compares all systems; pairs of numbers are \bench{}-Y/W.
CLIP, SigLIP2 and BGE-VL stay below 5 MRR on every split, and Qwen3-VL-Embedding-8B reaches only 8.2/14.0 MRR on known entities, retrieving look-alikes (\cref{fig:teaser}).
Generating the answer and matching it to gallery names gives 6.8/14.9 (16.7/23.5 with our graph context), as one string cannot rank the many answers of 1-to-N relations, and ComplEx falls from 36.2/37.5 with the oracle ID to 23.4/32.1 with our re-ID links.
On unseen entities, KG completion has only an untrained ID (ComplEx near zero, NativE 4.0/13.5 from image and text features), whereas \method{} starts from the photo: recall alone reaches 9.8/21.9 as the MLLM often names the entity (\cref{tab:abl}(d)), and listwise verification adds candidate facts.

\paragraph{Training on \bench{} teaches priors, reading teaches facts.}
Trained on our graph with our loss, the embedders improve sharply (CLIP4Cir Combiner 15.3/28.0, Qwen3-VL-Embedding 21.3/35.5 MRR on known entities).
Before listwise verification, \method{} leads on known entities by 61.5\%/14.9\% (34.4/40.8), but on unseen entities the fine-tuned embedder is ahead (12.9/27.2 vs.\ 11.9/24.2), where answers follow from relation statistics or appearance.

\paragraph{Reading the candidates closes the gap on unseen entities.}
With the listwise verifier, \method{} matches the fine-tuned embedder on unseen entities (14.2 vs.\ 12.9, 27.1 vs.\ 27.2) and leads by 70.0\%/24.2\% on known ones.
On the embedder's own top-20, the verifier reaches 15.7/30.9 on unseen entities, so most of the gain there comes from verification itself, while on known entities \method{}'s candidates add 32.1\%/6.5\%.
The full system reaches 39.1/46.7 MRR on known and 16.2/30.6 on unseen entities, the best MRR on five of the six test splits, 25.6\%/12.5\% above the fine-tuned embedder on unseen entities and 6.0\% above NativE with the oracle ID on known \bench{}-Y entities.

\subsection{In-the-wild photos: \bench{}-OVEN}
\label{sec:oven}
Before listwise verification, \method{} reaches 45.0 MRR on known \bench{}-OVEN entities, ahead of the fine-tuned embedder (42.3) and of NativE with the oracle ID (34.5).
Although only 38\% of the photos are re-identified correctly, reading the graph matters more than on web photos (removing it costs 11.5\% vs.\ 7.4\%/6.1\% MRR, 2-epoch ablations).
On unseen entities, \method{}'s own candidates fall short (29.2 vs.\ 32.5), so an appearance retriever proposes the answer and the verifier picks it out (37.2 on the embedder's candidates, 38.0 on the fused ones).

\subsection{Ablations}
\label{sec:abl}
\Cref{tab:abl}(a) isolates the design of recall on Qwen3.5-27B.
On known entities, fine-tuning matters most (stage 1 uses frozen states without the graph); removing name anchoring or graph reading costs 6.1--7.4\%, and graph reading also helps frozen states (21.4$\to$23.7 and 31.9$\to$34.8 MRR on known entities).
Training with oracle links hurts, because the model then over-trusts a context that is wrong for a third of the test photos, and extra link-dropout brings nothing.
Removing name verification of the top-50 costs 8.5--21.1\% MRR.

\subsection{Listwise verification}
\label{sec:exp_listwise}
On \bench{}-OVEN (\cref{tab:abl}(b)), removing the listwise verifier costs 16.6\%/18.4\% MRR on known/unseen entities, and the untuned backbone with the same prompt falls 12.7\%/16.3\% short.
The candidates' KG facts matter most and their photos help a little on unseen entities only; hiding the depicted entity trades 1.1\% on known for 2.4\% on unseen entities, and 20 candidates beat 10.
Applied directly to the fused priors, it reaches 57.3/38.0 with one forward pass instead of 50 (0.9\,s vs.\ 4.8\,s).

\subsection{Analysis}
In \cref{fig:analysis}, frozen answer-onset states carry more relational knowledge than describe or identity states, and only in late layers~(a).
Graph reading puts \method{} ahead of the fine-tuned embedder on known entities from the recall stage on; unseen entities need the appearance prior and listwise verification~(b).
The retrievers find different answers~(c), and \method{} wins three quarters of the relations of \bench{}-Y/W but under a third on \bench{}-OVEN~(d).
\method{} stays ahead on known entities at both gallery sizes~(e), and priors with listwise verification (1.5\,s per query) match the full system (6.3\,s)~(f).

Name verification adds 7.5--23.7\% MRR on both verified backbones, and Qwen3.5-27B is best on \bench{}-Y (\cref{tab:abl}(c)).
Split by the first hop (\cref{tab:abl}(d)), known entities with a correct link have 18.8--47.1\% higher MRR, yet \method{} stays useful when the link is wrong (38.2 on \bench{}-W), and on unseen entities whether the MLLM can name the photo matters most.
An oracle link at test time adds at most 7.0\% to recall MRR, so most errors lie in recall, not linking.

\subsection{Qualitative results}
In \cref{fig:qual_cases}, zero-shot generation and the fine-tuned embedder often return a famous or look-alike entity of the right type, while \method{} returns the related one.
Listwise verification fixes errors using the candidates' facts (Hollein replaces Gehry once the verifier reads that he died in Vienna); failures follow a wrong link to a look-alike: \method{} then returns an entity tied to it (Jeff Bridges of \emph{True Grit}) or a famous answer of the right type (Hitchcock), and listwise verification does not undo the link.

%% file: fig/main_analysis.tex
\begin{figure}[t]
\centering
\includegraphics[width=\linewidth]{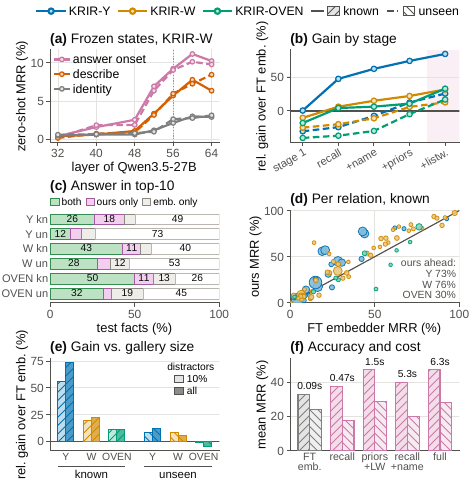}
\caption{\textbf{Analysis} (test MRR). (a)~Frozen states by layer (dotted: layer read). (b,\,e)~Relative gain over the fine-tuned embedder by stage, and with 10\% or all gallery distractors. (c)~Answers in the top-10 of both retrievers, one or neither. (d)~Per relation. (f)~Mean over benchmarks, with seconds per query; LW: listwise verification. (c--e)~Before listwise verification.}
\label{fig:analysis}
\end{figure}

%% file: tables/ablation_merged.tex
\begin{table}[!t]
\centering
\renewcommand{\arraystretch}{0.9}
\footnotesize
\setlength{\tabcolsep}{1.7pt}
\begin{tabular}{l cc cc}
\toprule
\multicolumn{5}{l}{\textit{(a) Recall design, test MRR (no priors, 2 epochs)}} \\
 & \multicolumn{2}{c}{\bench{}-Y} & \multicolumn{2}{c}{\bench{}-W} \\
\cmidrule(lr){2-3}\cmidrule(lr){4-5}
 & Known & Unseen & Known & Unseen \\
\midrule
\method{} ($\lambda{=}1$) + verif. & 31.9 & 12.8 & 40.1 & 25.8 \\
w/o verification & 28.3{\tiny\color{deltagray}\hspace{0.6pt}$-$11.3\%} & 10.1{\tiny\color{deltagray}\hspace{0.6pt}$-$21.1\%} & 36.7{\tiny\color{deltagray}\hspace{0.6pt}$-$8.5\%} & 22.1{\tiny\color{deltagray}\hspace{0.6pt}$-$14.3\%} \\
\midrule
Recall, $\lambda{=}0.5$ (ref.) & 28.2 & 9.7 & 36.1 & 22.1 \\
stage 1 (frozen, no graph) & 21.4{\tiny\color{deltagray}\hspace{0.6pt}$-$24.1\%} & 9.0{\tiny\color{deltagray}\hspace{0.6pt}$-$7.2\%} & 31.9{\tiny\color{deltagray}\hspace{0.6pt}$-$11.6\%} & 20.5{\tiny\color{deltagray}\hspace{0.6pt}$-$7.2\%} \\
w/o graph reading & 26.1{\tiny\color{deltagray}\hspace{0.6pt}$-$7.4\%} & 8.8{\tiny\color{deltagray}\hspace{0.6pt}$-$9.3\%} & 33.9{\tiny\color{deltagray}\hspace{0.6pt}$-$6.1\%} & 21.2{\tiny\color{deltagray}\hspace{0.6pt}$-$4.1\%} \\
w/o name anchoring ($\lambda{=}0$) & 26.3{\tiny\color{deltagray}\hspace{0.6pt}$-$6.7\%} & 9.0{\tiny\color{deltagray}\hspace{0.6pt}$-$7.2\%} & 33.7{\tiny\color{deltagray}\hspace{0.6pt}$-$6.6\%} & 20.1{\tiny\color{deltagray}\hspace{0.6pt}$-$9.0\%} \\
oracle links in training & 26.5{\tiny\color{deltagray}\hspace{0.6pt}$-$6.0\%} & 9.6{\tiny\color{deltagray}\hspace{0.6pt}$-$1.0\%} & 35.9{\tiny\color{deltagray}\hspace{0.6pt}$-$0.6\%} & 21.7{\tiny\color{deltagray}\hspace{0.6pt}$-$1.8\%} \\
link-dropout ($p{=}0.3$) & 27.7{\tiny\color{deltagray}\hspace{0.6pt}$-$1.8\%} & 9.6{\tiny\color{deltagray}\hspace{0.6pt}$-$1.0\%} & 35.5{\tiny\color{deltagray}\hspace{0.6pt}$-$1.7\%} & 21.6{\tiny\color{deltagray}\hspace{0.6pt}$-$2.3\%} \\
\bottomrule
\end{tabular}

\vspace{5pt}
\begin{tabular}{l cc cc}
\toprule
\multicolumn{5}{l}{\textit{(b) Listwise verifier on \bench{}-OVEN, test}} \\
 & \multicolumn{2}{c}{Known} & \multicolumn{2}{c}{Unseen} \\
\cmidrule(lr){2-3}\cmidrule(lr){4-5}
 & MRR & H@1 & MRR & H@1 \\
\midrule
No listwise verification & 46.8{\tiny\color{deltagray}\hspace{0.6pt}$-$16.6\%} & 35.6 & 31.0{\tiny\color{deltagray}\hspace{0.6pt}$-$18.4\%} & 21.8 \\
\midrule
Untuned backbone & 49.0{\tiny\color{deltagray}\hspace{0.6pt}$-$12.7\%} & 38.0 & 31.8{\tiny\color{deltagray}\hspace{0.6pt}$-$16.3\%} & 22.6 \\
Listwise verifier (ours) & 56.1 & 46.1 & 38.0 & 28.6 \\
w/o candidate photos & 56.2{\tiny\color{deltagray}\hspace{0.6pt}$+$0.2\%} & 46.7 & 37.3{\tiny\color{deltagray}\hspace{0.6pt}$-$1.8\%} & 27.9 \\
w/o candidate facts & 53.5{\tiny\color{deltagray}\hspace{0.6pt}$-$4.6\%} & 43.3 & 35.1{\tiny\color{deltagray}\hspace{0.6pt}$-$7.6\%} & 26.0 \\
w/o hiding the entity & 56.7{\tiny\color{deltagray}\hspace{0.6pt}$+$1.1\%} & 47.0 & 37.1{\tiny\color{deltagray}\hspace{0.6pt}$-$2.4\%} & 27.4 \\
$M{=}10$ candidates & 54.4{\tiny\color{deltagray}\hspace{0.6pt}$-$3.0\%} & 45.8 & 35.8{\tiny\color{deltagray}\hspace{0.6pt}$-$5.8\%} & 27.3 \\
\midrule
On priors, no name verif. & 57.3{\tiny\color{deltagray}\hspace{0.6pt}$+$2.1\%} & 48.2 & 38.0{\tiny\color{deltagray}\hspace{0.6pt}$\pm$0.0\%} & 29.0 \\
\bottomrule
\end{tabular}

\vspace{5pt}
\begin{tabular}{l cc cc}
\toprule
\multicolumn{5}{l}{\textit{(c) Backbones, test MRR (same 2-epoch recipe, $\lambda{=}0.5$)}} \\
 & \multicolumn{2}{c}{\bench{}-Y} & \multicolumn{2}{c}{\bench{}-W} \\
\cmidrule(lr){2-3}\cmidrule(lr){4-5}
 & Known & Unseen & Known & Unseen \\
\midrule
Qwen2.5-VL-7B$^\ast$, recall & 25.7 & 9.3 & 34.8 & 22.4 \\
\midrule
Qwen3-VL-32B, recall & 23.0 & 9.3 & 35.4 & 22.4 \\
Qwen3-VL-32B + verif. & 26.6{\tiny\color{deltagray}\hspace{0.6pt}$+$15.7\%} & 11.5{\tiny\color{deltagray}\hspace{0.6pt}$+$23.7\%} & 38.1{\tiny\color{deltagray}\hspace{0.6pt}$+$7.6\%} & 24.9{\tiny\color{deltagray}\hspace{0.6pt}$+$11.2\%} \\
\midrule
Qwen3.5-27B, recall & 28.2 & 9.7 & 36.1 & 22.1 \\
Qwen3.5-27B + verif. & 31.4{\tiny\color{deltagray}\hspace{0.6pt}$+$11.3\%} & 11.9{\tiny\color{deltagray}\hspace{0.6pt}$+$22.7\%} & 38.8{\tiny\color{deltagray}\hspace{0.6pt}$+$7.5\%} & 24.4{\tiny\color{deltagray}\hspace{0.6pt}$+$10.4\%} \\
\bottomrule
\end{tabular}

\vspace{5pt}
\begin{tabular}{l cc cc}
\toprule
\multicolumn{5}{l}{\textit{(d) Errors by the first hop, test MRR (share of facts)}} \\
 & \multicolumn{2}{c}{Known: link} & \multicolumn{2}{c}{Unseen: MLLM names it} \\
\cmidrule(lr){2-3}\cmidrule(lr){4-5}
 & correct & wrong & yes & no \\
\midrule
\bench{}-Y & 42.5{\tiny\color{deltagray}\hspace{0.6pt}59\%} & 28.9{\tiny\color{deltagray}\hspace{0.6pt}41\%} & 25.4{\tiny\color{deltagray}\hspace{0.6pt}33\%} & 9.1{\tiny\color{deltagray}\hspace{0.6pt}67\%} \\
\bench{}-W & 45.4{\tiny\color{deltagray}\hspace{0.6pt}71\%} & 38.2{\tiny\color{deltagray}\hspace{0.6pt}29\%} & 39.7{\tiny\color{deltagray}\hspace{0.6pt}41\%} & 21.3{\tiny\color{deltagray}\hspace{0.6pt}59\%} \\
\bench{}-OVEN & 55.1{\tiny\color{deltagray}\hspace{0.6pt}35\%} & 42.3{\tiny\color{deltagray}\hspace{0.6pt}65\%} & 47.9{\tiny\color{deltagray}\hspace{0.6pt}24\%} & 25.7{\tiny\color{deltagray}\hspace{0.6pt}76\%} \\
\bottomrule
\end{tabular}
\caption{\textbf{Ablations and analysis} (\%). Grey in (a--c): relative change of MRR with respect to the reference row (the first row of each block in (a), the listwise verifier in (b), recall with the same backbone in (c)); in (b), the last row re-ranks the fused priors directly with one forward pass instead of 50. (a)~Name verification of the top-50. (c)~Name verification of the top-20; $^\ast$trained with link-dropout. (d)~Before listwise verification; known: whether the linker's top-1 is the depicted entity; unseen: whether Qwen3.5-27B names it; grey: share of facts.}
\label{tab:abl}
\end{table}

%% file: fig/qual_cases.tex
\begin{figure*}[t]
\centering
\qcase{Who is this person married to?}{\includegraphics[width=\linewidth]{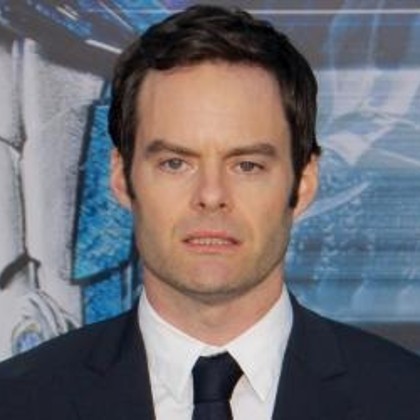}}{%
\qa{ZS}{Olivia Wilde\cno}\qa{FT}{Patricia Arquette\cno}\qa{Ours}{Maggie Carey\cok}}{\bench{}-Y, known; linked to Bill Hader\cok}\hfill
\qcase{Which studio made this film?}{\includegraphics[width=\linewidth]{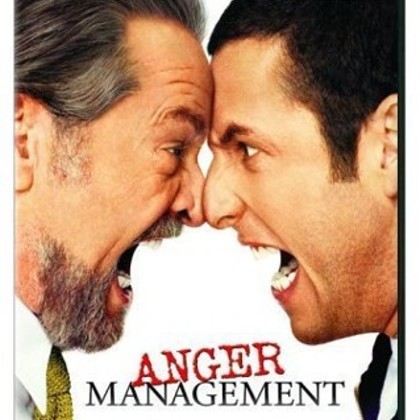}}{%
\qa{ZS}{Columbia Pictures\cno}\qa{FT}{Columbia Pictures\cno}\qa{Ours}{Happy Madison Productions\smash{$^\dagger$}\cok}}{\bench{}-W, known; linked to \emph{Anger Management}\cok}\hfill
\qcase{Who designed this building?}{\includegraphics[width=\linewidth]{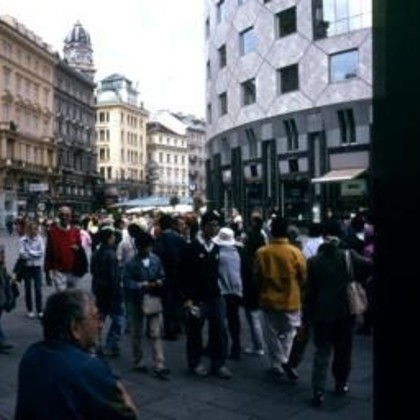}}{%
\qa{ZS}{Frank Gehry\cno}\qa{FT}{Frank Gehry\cno}\qa{Ours}{Hans Hollein\smash{$^\dagger$}\cok}}{\bench{}-OVEN, known; linked to Dancing House (Gehry)\cno}\hfill
\qcase{Which country is this beach in?}{\includegraphics[width=\linewidth]{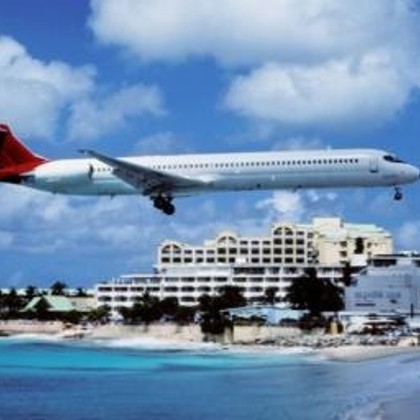}}{%
\qa{ZS}{United States\cno}\qa{FT}{United States\cno}\qa{Ours}{Kingdom of the Netherlands\cok}}{\bench{}-OVEN, known; linked to Maho Beach\cok}
\\[2pt]
\qcase{Who directed this film?}{\includegraphics[width=\linewidth]{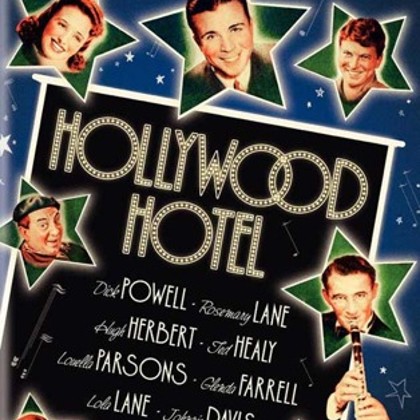}}{%
\qa{ZS}{Walter Lang\cno}\qa{FT}{Roy Del Ruth\cno}\qa{Ours}{Busby Berkeley\cok}}{\bench{}-Y, unseen; linked to \emph{Destiny} (1942)}\hfill
\qcase{What is this person's job?}{\includegraphics[width=\linewidth]{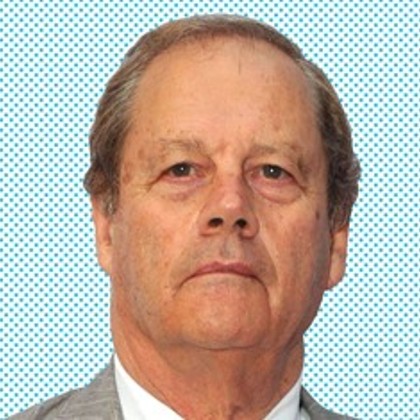}}{%
\qa{ZS}{Actor\cno}\qa{FT}{Lawyer\cno}\qa{Ours}{Film director\cok}}{\bench{}-W, unseen; linked to Stuart Gillard, a film director}\hfill
\qcase{Which city is this stadium in?}{\includegraphics[width=\linewidth]{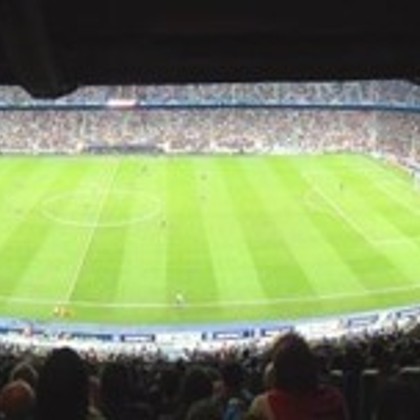}}{%
\qa{ZS}{London\cno}\qa{FT}{Madrid\cno}\qa{Ours}{Barcelona\smash{$^\dagger$}\cok}}{\bench{}-OVEN, unseen; linked to City of Manchester Stadium}\hfill
\qcase{Who was born in this castle?}{\includegraphics[width=\linewidth]{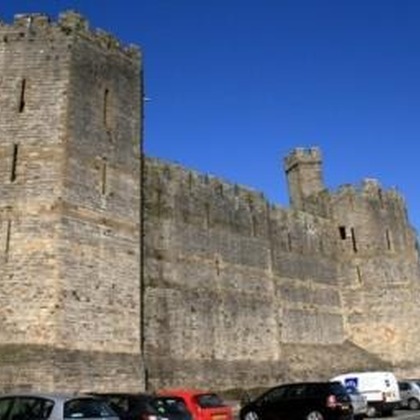}}{%
\qa{ZS}{Richard, 3rd Earl of Cornwall\cno}\qa{FT}{John Hastings, 2nd Earl of Pembroke\cno}\qa{Ours}{Edward II of England\smash{$^\dagger$}\cok}}{\bench{}-OVEN, unseen; linked to Portchester Castle}
\\[2pt]
\qcase{Who acted in this film?}{\includegraphics[width=\linewidth]{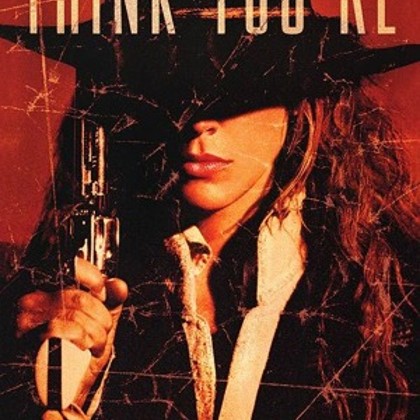}}{%
\qa{ZS}{Uma Thurman\cno}\qa{FT}{Liam Neeson\cno}\qa{Ours}{Jeff Bridges\smash{$^\dagger$}\cno}\qa{GT}{Gene Hackman}}{\bench{}-W, known; linked to \emph{True Grit} (2010)\cno}\hfill
\qcase{What does this dish contain?}{\includegraphics[width=\linewidth]{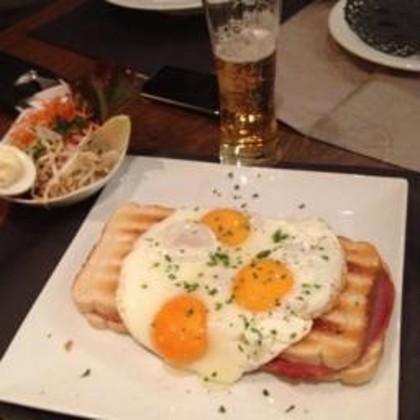}}{%
\qa{ZS}{Egg\cno}\qa{FT}{Yolk\cno}\qa{Ours}{Yolk\smash{$^\dagger$}\cno}\qa{GT}{Ham}}{\bench{}-OVEN, known; linked to Eggs Benedict\cno}\hfill
\qcase{Who directed this film?}{\includegraphics[width=\linewidth]{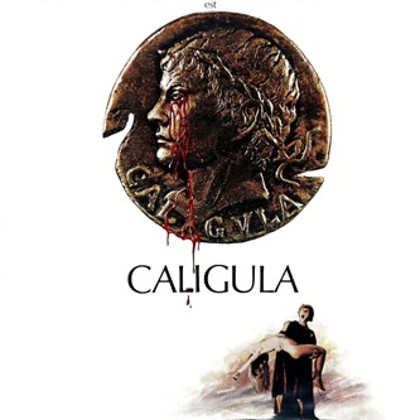}}{%
\qa{ZS}{Tinto Brass\cok}\qa{FT}{Malcolm McDowell\cno}\qa{Ours}{Ken Russell\smash{$^\dagger$}\cno}\qa{GT}{Tinto Brass}}{\bench{}-Y, known; linked to \emph{Julius Caesar} (1970)\cno}\hfill
\qcase{Who directed this film?}{\includegraphics[width=\linewidth]{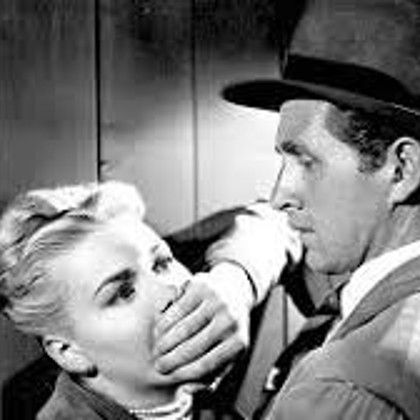}}{%
\qa{ZS}{Alfred Hitchcock\cno}\qa{FT}{Harold D. Schuster\cno}\qa{Ours}{Alfred Hitchcock\smash{$^\dagger$}\cno}\qa{GT}{Edward Dmytryk}}{\bench{}-Y, known; linked to \emph{Working Girls} (1931)\cno}
\caption{\textbf{Qualitative results} for known (top) and unseen (middle) entities, and failures (bottom). ZS: Qwen3.5-27B zero-shot; FT: fine-tuned embedder; Ours: \method{} ($\dagger$: changed by listwise verification); GT: correct answer; grey: top-1 linked entity, with \cmark/\xmark{} for a correct/wrong link on known entities. Questions are paraphrased.}
\label{fig:qual_cases}
\end{figure*}

%% file: sec/6_conclusion.tex
\section{Conclusion}
\label{sec:conclusion}
We introduced relational image retrieval over multimodal knowledge graphs and three benchmarks for it, one with photos taken in the wild.
\method{} lets one MLLM use the MMKG at both ends of a query, the re-identified entity's facts to recall the answer and the candidates' own facts and photos to verify it, and the full system beats a fine-tuned embedder on known and unseen entities of every benchmark.

\paragraph{Limitations.}
\bench{} inherits the noise of its MMKGs, and unseen in-the-wild entities still need an appearance retriever to propose candidates.
\par

%% file: sec/appendix.tex
\begin{table*}[h]
\centering
\small
\begin{tabularx}{\textwidth}{lX}
\toprule
Backbone & Qwen3.5-27B, bf16, thinking disabled \\
Image resolution & $\le 448^2$ pixels \\
LoRA & $r=16$, $\alpha=32$, dropout 0.05; q/k/v/o, in\_proj\_qkv/z, out\_proj, gate/up/down \\
Heads & LayerNorm + Linear $\to$ 1024, learnable temperature \\
Stage 1 & frozen states (layer 56), 15 epochs, AdamW $10^{-3}$, batch 512 \\
Stage 2 & AdamW, LR $10^{-4}$ (LoRA) / $2{\times}10^{-4}$ (heads), cosine, 3\% warm-up \\
Batch / epochs & 64 queries (8 A800 $\times$ 8) / 4, 6, 10 on \bench{}-Y/W/OVEN \\
Loss & full-gallery multi-positive InfoNCE + $1.0\,\mathcal{L}_{\text{name}}$ \\
Training links & real linker output on training photos (top-3), $\le$12 facts; no link-dropout \\
Query state & layer 56, answer onset (after the empty think block) \\
Entity memory & frozen, layer 56, last name token \\
Linker & CLIP-L/14-336 $\oplus$ centred identity state (layer 56) \\
Verification & top-50, summed name log-likelihood, weight $\eta$ on validation \\
Priors & $T,\beta,\gamma$ chosen on validation MRR \\
Listwise verifier & separate LoRA ($r=16$), $M{=}20$, $256^2$ photo + 3 facts per candidate, hiding prob.\ 0.3, \\
 & 1/1/2 epochs on \bench{}-Y/W/OVEN (12k queries on \bench{}-Y), batch 32, LR $10^{-4}$ \\
\bottomrule
\end{tabularx}
\caption{Hyper-parameters of \method{}.}
\label{tab:hp}
\end{table*}

\input{tables/main}

\input{tables/oven}

\input{tables/seeds}

\begin{figure*}[t]
\centering
\includegraphics[width=\linewidth]{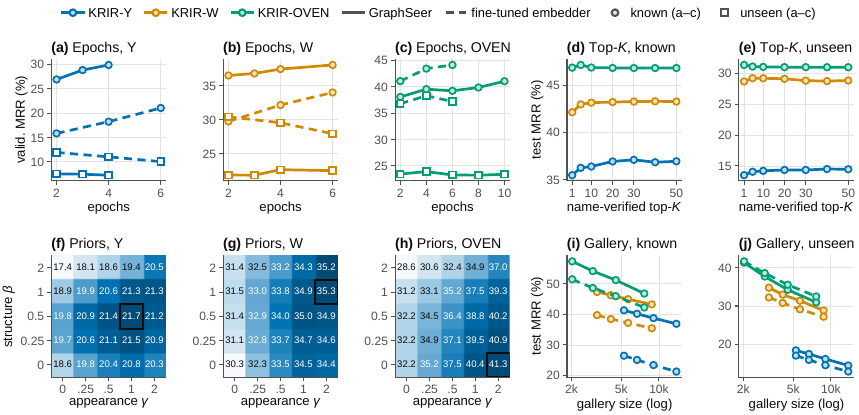}
\caption{\textbf{Training length and sensitivity.} (a--c)~Validation MRR against training epochs for \method{} recall (solid) and the fine-tuned embedder (dashed); circles: known, squares: unseen entities. (d,e)~Test MRR before listwise verification against the number $K$ of name-verified candidates (weight chosen on validation for each $K$). (f--h)~Mean validation MRR over the structural ($\beta$) and appearance ($\gamma$) prior weights at the selected temperature; box: best cell. (i,j)~Test MRR against gallery size (random distractor subsets, all answers kept; mean of three draws), before listwise verification; dashed: fine-tuned embedder.}
\label{fig:appx_training}
\end{figure*}

\input{tables/sweep}

\input{tables/verifyk}

\section{Benchmarks}
\label{sec:appx_bench}
\paragraph{Image extraction.}
We decode every image in the MMRNS~\cite{xu2022mmrns} archives, convert it to RGB, and drop images smaller than $32$ px.
Each remaining image is resized to a longest side of 448 px.
Images are matched to entities by normalised folder names; on MKG-W this goes through the DBpedia$\leftrightarrow$Wikidata links shipped with the dataset.
In total, 14{,}480 YAGO entities have 43{,}346 images, and 9{,}278 Wikidata entities have 27{,}814 images.
\paragraph{Query photos.}
For each entity, the query photo is the first non-gallery image whose 256-bit difference hash differs from the gallery photo's hash by more than 24 bits.
14{,}411 YAGO and 9{,}250 Wikidata entities have such a photo.
\paragraph{Splits.}
The unseen-entity pool consists of entities that have a query photo and head at least one validation or test fact.
10\% of this pool (seed 0) is removed from the training graph and from the gallery.
One third of the removed entities serve inductive validation and two thirds inductive test.
A removed entity's facts in the original train, validation and test splits become its inductive queries, keeping those whose answer is still in the gallery.
Training triples touching a removed entity are dropped; there are 19{,}238 and 31{,}381 remaining training triples.
\paragraph{Filtering.}
Ranks are filtered with the answers of the original train, validation and test splits.
Each fact contributes one rank, and ties count half.

\paragraph{\bench{}-OVEN construction.}
We use the entity split of OVEN~\cite{hu2023oven} and exclude its VQA-derived query split, whose labels answer a question about the photo rather than name what it shows.
Gallery photos are the entities' Wikipedia images from the M-BEIR release~\cite{wei2024uniir}; an entity without one is represented by one of its OVEN photos, never by the query photo.
From Wikidata5M~\cite{wang2021kepler} we drop bookkeeping, taxonomic-rank and class-membership properties, keep relations with at least 30 facts, and keep $(h,r)$ groups with at most 20 answers.
A fact is queryable if its head has an OVEN photo and its answer has an image; all other Wikidata5M facts among the selected entities enter the training graph as readable context.
Queryable facts are split 80/10/10, and the regimes are formed as for \bench{}-Y/W, with 30\% of the eligible entities unseen so that the smaller benchmark keeps enough inductive queries.

\paragraph{\bench{}-OVEN photo check.}
We ask Qwen3.5-27B ``Does this image show $\langle$name$\rangle$?'' for every validation and test query photo and read $p(\text{yes})$ from the first answer token.
89\% of the photos pass ($p>0.5$); the rest are mostly lesser-known landmarks and species that the model cannot confirm, plus a few generic concepts (\eg ``Smoking'').
On the verified test subset (550/615 transductive and 584/659 inductive queries), every system improves slightly and the ranking stays the same, with 43.0/33.8 MRR for the fine-tuned embedder, 45.1/23.9 for \method{} and 47.6/32.4 for \method{} with priors (both before listwise verification).

\section{Implementation details}
\label{sec:appx_impl}
\paragraph{Baselines.}
MLLM embedders receive the photo and the relation question under their own retrieval instruction, against photo or photo+name galleries, whichever is stronger on validation.
The CLIP4Cir Combiner~\cite{baldrati2022clip4cir} fuses frozen CLIP-L/14 photo and question features.
The fine-tuned Qwen3-VL-Embedding-8B uses LoRA against its own frozen photo+name gallery; learning rate, rank, temperature and epochs are tuned on validation with our tuning budget.
The untuned name verifier re-scores the embedder's top-20 by Qwen3.5-27B name likelihoods given the same graph context.
The generative baseline answers with greedy decoding plus 3 samples; answers are matched to gallery names with Qwen3-Embedding and averaged over samples.
ComplEx-N3~\cite{trouillon2016complex,lacroix2018n3} is trained on the training graph and uses the soft re-ID linking of our structural prior.
NativE~\cite{zhang2024native} uses the authors' entity features on \bench{}-Y/W and, on \bench{}-OVEN, which has none, CLIP-L/14 features of each entity's photo and name, with the default hyper-parameters on all three.
Without fine-tuning, we match frozen answer-onset states zero-shot or through a linear head.

\paragraph{Hyper-parameters.}
\Cref{tab:hp} lists the settings of \method{}.

\paragraph{Prompts.}
All prompts use the chat template of the backbone with thinking disabled, and the image is placed before the text.
\begin{itemize}\itemsep0pt
\item \textbf{Identity (gallery memory, linker):} ``What is the name of the specific entity (person, place, organization, creative work, etc.) shown in this image? Answer with the name only.'' For the entity memory, the KG name is appended as the assistant answer, and the state at its last token is read.
\item \textbf{Relational recall (query):} \texttt{[context]} + the relation template with $x=$``the entity shown in the image'' + `` Answer with the name only.''
\item \textbf{Context:} ``Knowledge-graph lookup by visual match (score): $n_1$ ($s_1$), $n_2$ ($s_2$), $n_3$ ($s_3$). Known facts about $n_1$:'' followed by up to 12 lines ``- \texttt{template}(it) \texttt{answer}''. Facts of the queried relation are removed.
\item \textbf{Describe control (E5-V~\cite{jiang2024e5v} style):} query ``\texttt{question} Summarize the answer in one word:''; gallery ``Summarize above image in one word:''.
\end{itemize}

\paragraph{Relation templates.}
YAGO relations use hand-written forward and inverse questions, \eg \texttt{wasBornIn}: ``Where was $x$ born?'' / ``Who was born in $x$?'', and \texttt{playsFor}: ``Which team does $x$ play for?'' / ``Who plays for $x$?''.
Wikidata relations use their official English labels in the templates ``What is the \texttt{label} of $x$?'' and ``Which entity has $x$ as its \texttt{label}?''.
The full list will be released with the code.

\section{Additional results}
\label{sec:appx_results}
\paragraph{Full results.}
\Cref{tab:main_full,tab:oven_full} list all four metrics (MRR, H@1, H@3, H@10) for the systems of \cref{tab:main}.

\paragraph{Seeds.}
\Cref{tab:seeds} repeats stage 2 of the final configuration with three seeds; the standard deviation is at most 0.6 MRR.
A second seed of the listwise verifier gives 39.1/16.2 (\bench{}-Y), 46.7/30.4 (\bench{}-W) and 56.6/36.9 (\bench{}-OVEN), against 39.1/16.2, 46.7/30.6 and 56.1/38.0 for the reported seed.

\begin{figure}[h]
\centering
\includegraphics[width=\linewidth]{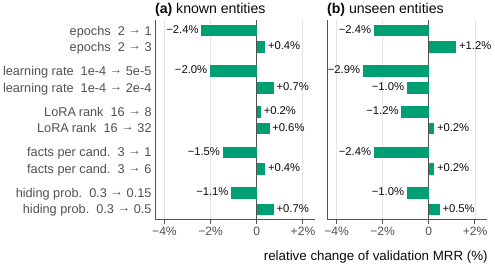}
\caption{\textbf{Listwise verifier hyper-parameters} on \bench{}-OVEN: relative change of validation MRR on known (a) and unseen (b) entities when one setting of the recipe is changed (default $\to$ new value).}
\label{fig:appx_lw_sens}
\end{figure}

\paragraph{Hyper-parameters.}
\Cref{fig:appx_training}(a--c) shows validation MRR against the number of training epochs, swept with the same budget for \method{} and the fine-tuned embedder.
Longer training helps both on known entities, while the embedder loses on unseen entities as it fits the training graph and \method{} stays flat there; we select the epochs per benchmark and model on validation (\method{}: 4/6/10 on \bench{}-Y/W/OVEN; embedder: 6/6/4).
\Cref{tab:sweep} varies the LoRA rank, the name-anchoring weight, the learning rate and the number of epochs on \bench{}-W in a matched 2-GPU setting; a larger name-anchoring weight ($\lambda{=}1$) helps most and was adopted after confirming it on validation of both datasets.
For the listwise verifier on \bench{}-OVEN (\cref{fig:appx_lw_sens}), seven of ten single-factor changes move validation MRR by at most 1.2\% on known and unseen entities; fewer updates (1 epoch, or learning rate $5{\times}10^{-5}$) and a single fact per candidate cost 1.5--2.9\%.
Training longer on \bench{}-W, training on all \bench{}-Y queries, showing 26 candidates at test time or averaging over the reversed candidate order changes MRR by at most 0.4, so we keep the recipe fixed.

\begin{figure*}[t]
\centering
\includegraphics[width=\linewidth]{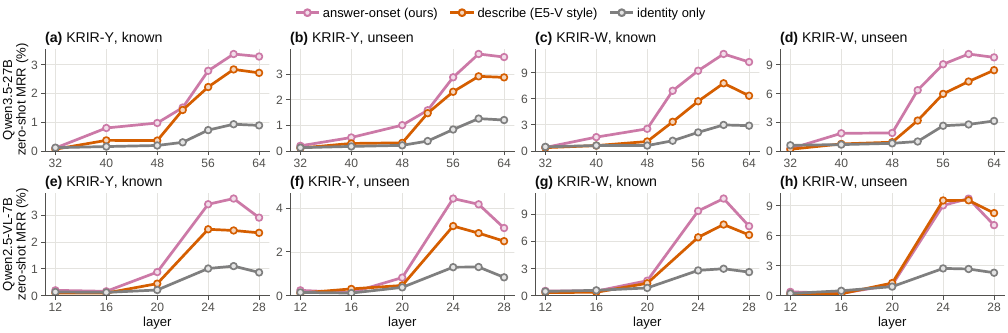}
\caption{\textbf{Zero-shot retrieval with frozen MLLM states} by layer (test MRR): Qwen3.5-27B (a--d) and Qwen2.5-VL-7B (e--h). Answer-onset states, read when the model is about to name the answer, carry more relational knowledge than describe-style or identity-only states, and this knowledge appears late.}
\label{fig:layers}
\end{figure*}

\begin{figure*}[h]
\centering
\includegraphics[width=\linewidth]{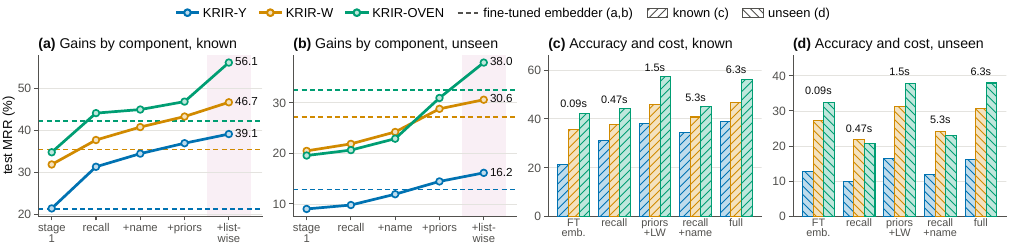}
\caption{\textbf{Gains and cost} (test MRR). (a,b)~\method{} from frozen states (stage 1) through recall, name verification and the priors to listwise verification (the final system, shaded); dashed: the Qwen3-VL-Embedding-8B fine-tuned with the same data and loss, in the colour of its benchmark. (c,d)~MRR of each benchmark for the systems sorted by online cost per query (batch 1, one A800, measured on \bench{}-Y); LW: listwise verification. Systems with priors run both models.}
\label{fig:appx_pipeline}
\end{figure*}

\begin{figure*}[h]
\centering
\includegraphics[width=\linewidth]{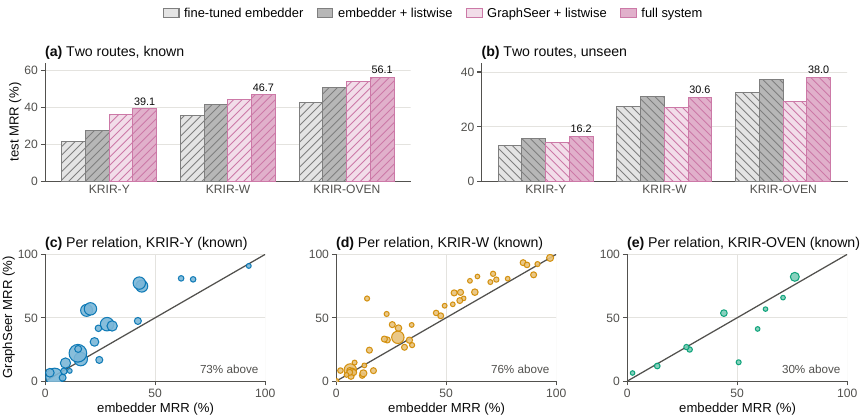}
\caption{\textbf{Two routes to the answer} (test MRR). (a,b)~The fine-tuned embedder, its own top-20 re-ranked by the listwise verifier, \method{}'s candidates re-ranked by it, and the full system (value given). (c--e)~Per-relation MRR on known entities, before listwise verification; one point per relation with $\ge$15 test facts, area proportional to the number of facts.}
\label{fig:appx_routes}
\end{figure*}

\begin{figure*}[t]
\centering
\includegraphics[width=0.7\linewidth]{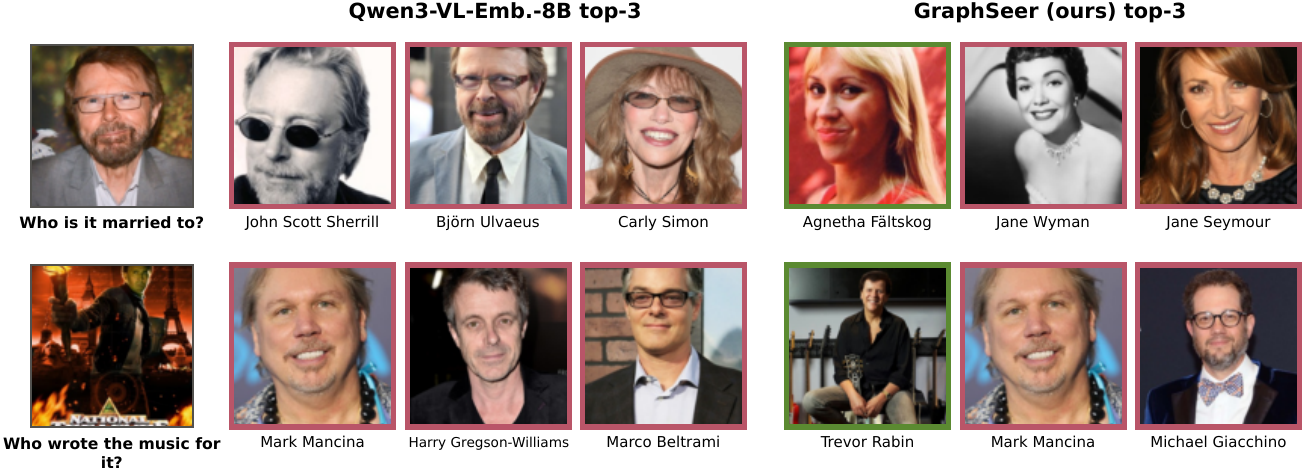}
\caption{\textbf{Look-alikes vs.\ related entities} (\bench{}-Y test): top-3 of the zero-shot Qwen3-VL-Embedding-8B and of \method{}'s recall stage. Green: correct; red: wrong.}
\label{fig:appx_teaser}
\end{figure*}

\paragraph{Verification budget.}
\Cref{tab:verifyk} varies the number $K$ of name-verified candidates and the token aggregation, without priors.
Gains grow with $K$ up to 50 and the summed log-likelihood is slightly better on validation, so the final model uses the top-50 with summation; the top-10 already recovers 57--74\% of the gain at a fifth of the cost.
With priors, gains saturate after $K\approx10$--$30$ (\cref{fig:appx_training}(d,e)).

\section{Analysis}
\label{sec:appx_analysis}
Numbers are test MRR (\%) unless noted, and differences are relative; \cref{fig:analysis}(c), \cref{fig:appx_routes}(c--e), \cref{fig:appx_training}(d--j) and \cref{tab:abl}(d) use the systems before listwise verification.

\subsection{Where relational knowledge lives}
Matching \emph{frozen} Qwen3.5-27B states zero-shot, answer-onset states beat E5-V-style describe and identity states from layer 40 on, on all four splits (11.1 vs.\ 7.8 MRR on known \bench{}-W entities at the best layers).
The signal is weak before layer 48 and peaks at layer 60 of 64, mirroring the late formation of visual subject representations~\cite{venhoff2025toolate} (\cref{fig:layers}(a--d)).
Qwen2.5-VL-7B shows the same ordering, except that describe states are on par on unseen \bench{}-W entities, and peaks at layers 24--26 of 28, \ie at the same relative depth (\cref{fig:layers}(e--h)).
The state also encodes which kind of entity can answer: a relation-range prior $\log p(t\mid r)$ estimated from training tails, a classic aid for link prediction, brings no gain, and validation selects a zero weight on both datasets.

\subsection{Gains, cost and the two routes}
\Cref{fig:appx_pipeline}(a,b) follows the final system step by step.
On known entities, fine-tuning with graph reading (stage 1$\to$recall) gives the largest single gain on \bench{}-Y/W and already passes the fine-tuned embedder on all three benchmarks.
On unseen entities, recall and name verification stay below the embedder; the appearance prior closes the gap on \bench{}-Y/W, and listwise verification puts all three benchmarks above it, by the widest margin on \bench{}-Y (16.2 vs.\ 12.9, $+$25.6\%).
On one A800 at batch 1 (\cref{fig:appx_pipeline}(c,d)), recall, including linking, takes 0.47\,s per query, five times an embedder query (0.09\,s), and name verification of the top-10/top-50 adds 1.0/4.8\,s.
The listwise verifier reads the top-20 in one forward pass of about 2.3k tokens (0.92\,s); applied to the fused priors, it needs 1.5\,s per query with both retrievers and stays within 2.6\% of the full system's MRR (6.3\,s) on every split, ahead of it on three of the six.
Of the two routes (\cref{fig:appx_routes}(a,b)), \method{}'s candidates add 6.1\% MRR on known \bench{}-OVEN entities over verifying the embedder's own candidates (32.1\%/6.5\% on \bench{}-Y/W); on unseen entities, verifying the embedder's candidates already gives most of the gain, and the full system adds a little on two of the three benchmarks.

\subsection{Knowledge versus appearance}
\label{sec:analysis}
The two retrievers find different answers (\cref{fig:analysis}(c)).
On known entities of \bench{}-Y/W, the facts whose answer only \method{} ranks in its top-10 (18\%/11\%) far outnumber those found only by the embedder (7\%/6\%); on \bench{}-OVEN they are similar (11\% vs.\ 13\%), and on unseen entities the embedder finds more.
Taking the better of the two for each fact would give 39.0/17.3 MRR on \bench{}-Y (known/unseen), against 34.4/12.9 for the stronger one alone; the full system, which verifies the candidates of both, reaches 39.1/16.2.
Per relation (\cref{fig:appx_routes}(c--e), \cref{tab:rel}), \method{} is ahead on 73\% and 76\% of the relations of \bench{}-Y and \bench{}-W, often by more than 30 MRR: for a footballer's team or a person's affiliation, the fine-tuned embedder reaches 43.9 and 42.8 and \method{}, which reads the linked entity's facts, 74.9 and 77.3.
Fine-tuning teaches the embedder relation-level priors (an unseen person's occupation rises from 2.0 to 93.6 MRR, as most people share a few occupations), so it keeps the advantage on relations whose answers are statistically predictable, such as the films in a given language (15.7 vs.\ 7.3).
On \bench{}-OVEN, \method{} wins the two largest, geographic relations (country 82.3 vs.\ 76.2, located-in 53.7 vs.\ 44.0), while the embedder wins most taxonomic and material relations, whose answers follow from appearance.

\begin{table}[h]
\centering
\small
\setlength{\tabcolsep}{3pt}
\resizebox{\linewidth}{!}{%
\begin{tabular}{llrrrr}
\toprule
 & & & \multicolumn{2}{c}{Emb.} & \\
\cmidrule(lr){4-5}
 & Relation (query $\to$ answer) & \#facts & ZS & FT & \method{} \\
\midrule
\multirow{3}{*}{\rotatebox{90}{Y, trans.}} & Which team does X play for? & 250 & 2.7 & 43.9 & \textbf{74.9} \\
 & Which organization is X affiliated to? & 270 & 8.3 & 42.8 & \textbf{77.3} \\
 & Who acted in X? (poster $\to$ actor) & 576 & 16.1 & 14.9 & \textbf{22.0} \\
\midrule
\multirow{3}{*}{\rotatebox{90}{W, ind.}} & occupation of X & 26 & 2.0 & 93.6 & \textbf{94.2} \\
 & cast member of X (poster $\to$ actor) & 133 & 20.8 & 24.1 & \textbf{24.4} \\
 & films whose original language is X & 225 & 3.0 & \textbf{15.7} & 7.3 \\
\bottomrule
\end{tabular}}
\caption{\textbf{Per-relation test MRR} of Qwen3-VL-Embedding-8B, zero-shot (ZS) and fine-tuned (FT), and of \method{} (recall + name verification).}
\label{tab:rel}
\end{table}

\subsection{Linking and reading}
\label{sec:appx_link}
\paragraph{Re-identification.}
Linking the held-out query photo to its own KG entity is far from solved.
Top-1 accuracy (\bench{}-Y\,/\,\bench{}-W) is 49.8\%\,/\,64.5\% with SigLIP2, 57.4\%\,/\,68.1\% with CLIP-L, and 53.9\%\,/\,65.3\% with Qwen3.5-27B identity states.
Concatenating CLIP and the identity states works best, with 60.7\%\,/\,71.2\% (37.6\% on \bench{}-OVEN), with an AUROC of 0.73\,/\,0.78 for separating seen from unseen entities by their top similarity.

\paragraph{Selective use of the graph.}
With the 2-epoch models of \cref{tab:abl}(a) ($\lambda{=}0.5$), we split known-entity test facts by whether the linker's top-1 is the query entity.
When the link is correct, reading the graph adds 11.4\%/8.5\% MRR on \bench{}-Y/W (28.3 vs.\ 25.4 and 37.2 vs.\ 34.3 without context); when it is wrong, the context still does not hurt (it adds 4.1\%/1.5\%).
The model trusts the graph when the match is right and otherwise relies on its own knowledge.

\paragraph{Upper bound from linking.}
With the 2-epoch models ($\lambda{=}1$), placing the true entity first among the linked candidates at test time (an oracle link), without retraining, raises recall MRR on known entities from 28.3/36.7/40.1 to 29.0/37.2/42.9 on \bench{}-Y/W/OVEN; unseen entities are unaffected because they are not in the gallery.
Naming-assisted linking, which moves gallery entities whose name matches the MLLM's own name for the photo to the top, raises top-1 re-identification by 5.7--9.1\% but leaves MRR unchanged (within 1.2\%), so better linking alone has limited headroom.

\paragraph{Counterfactual facts.}
We replace a true answer that appears in the read facts of a correctly linked known entity by another entity from the same relation's range (927/429 test and validation queries on \bench{}-Y/W).
\method{}'s top-1 stays the original answer for 74/83\% of them (82/85\% before the edit), and the replacement reaches rank 1 for at most 0.2\%, so recall weighs a single contradicting fact against what the model knows about the entity.
Moving the facts that reach the linked entity from the answer to the strongest wrong candidate raises that candidate's listwise probability from 0.02 to 0.11/0.12, while the verifier still prefers the answer in most cases.

\subsection{Sensitivity}
\Cref{fig:appx_training}(f--j) varies the remaining test-time choices of the final systems, with all other settings at their validation optimum.
The fusion weights form a broad plateau around the selected values; on \bench{}-W/OVEN the best appearance weight lies at the edge of the grid.
Removing random distractors from the gallery raises every method, and \method{} stays ahead of the fine-tuned embedder on known entities at every gallery size.
\section{Qualitative results}
\label{sec:appx_qual}
On two \bench{}-Y test queries (\cref{fig:appx_teaser}), the zero-shot embedder returns the query person himself and look-alikes for the spouse, and composers of similar films for the composer, while \method{}'s recall ranks the spouse and the composer first.

\paragraph{Failures.}
All four failures in \cref{fig:qual_cases} are known entities whose photo is linked to a look-alike of the same type: another western (\emph{True Grit}, 2010), another dish (eggs Benedict), another historical film (\emph{Julius Caesar}, 1970) and another black-and-white film (\emph{Working Girls}, 1931).
The wrong link misleads in two ways.
The answer may follow the look-alike's facts: Jeff Bridges is in the cast of \emph{True Grit}, and for the croque monsieur \method{} answers hollandaise sauce, a part of eggs Benedict, before listwise verification and yolk, a part of hollandaise, after it.
When the look-alike's facts do not apply, the model falls back on a well-known director, Alfred Hitchcock or Ken Russell.
Listwise verification changes the top-1 in all four cases but does not recover the answer (its rank moves 7$\to$15, 73$\to$73, 3$\to$6 and 14$\to$11, in the order of the figure); for \emph{The Quick and the Dead} it even follows the link, because candidate facts that mention a linked entity are shown first.
For \emph{Caligula}, zero-shot generation names the director from the title on the poster, so a wrong link can override what the backbone already knows.

%% file: tables/main.tex
\begin{table*}[t]
\centering
\small
\setlength{\tabcolsep}{3.6pt}
\resizebox{\textwidth}{!}{%
\begin{tabular}{l cccc cccc cccc cccc}
\toprule
 & \multicolumn{8}{c}{\bench{}-Y (YAGO)} & \multicolumn{8}{c}{\bench{}-W (Wikidata)} \\
\cmidrule(lr){2-9}\cmidrule(lr){10-17}
 & \multicolumn{4}{c}{Transductive} & \multicolumn{4}{c}{Inductive} & \multicolumn{4}{c}{Transductive} & \multicolumn{4}{c}{Inductive} \\
Method & MRR & H@1 & H@3 & H@10 & MRR & H@1 & H@3 & H@10 & MRR & H@1 & H@3 & H@10 & MRR & H@1 & H@3 & H@10 \\
\midrule
\tgroup{17}{Zero-shot embedders} \\
CLIP-L/14~\cite{radford2021clip} (image + relation text) & 1.0 & 0.2 & 1.1 & 2.1 & 1.0 & 0.6 & 1.0 & 1.8 & 2.4 & 0.6 & 3.2 & 5.2 & 2.9 & 1.6 & 3.0 & 5.1 \\
SigLIP2-so400m~\cite{tschannen2025siglip2} (image + text) & 0.7 & 0.1 & 0.7 & 1.4 & 0.6 & 0.2 & 0.6 & 0.9 & 2.3 & 0.9 & 2.7 & 4.2 & 2.3 & 1.1 & 2.7 & 3.9 \\
BGE-VL-L (MegaPairs)~\cite{zhou2025megapairs} & 0.3 & 0.0 & 0.0 & 0.4 & 0.4 & 0.1 & 0.3 & 0.6 & 0.9 & 0.3 & 0.7 & 1.6 & 0.8 & 0.3 & 0.4 & 1.5 \\
GME-Qwen2-VL-7B~\cite{zhang2025gme} & 5.7 & 1.4 & 7.2 & 12.3 & 6.9 & 4.0 & 7.2 & 11.9 & 12.2 & 4.5 & 16.4 & 24.7 & 13.4 & 8.9 & 14.5 & 22.1 \\
Qwen3-VL-Embedding-8B~\cite{li2026qwen3vlemb} & 8.2 & 4.0 & 9.5 & 14.9 & 10.3 & 7.3 & 10.4 & 15.4 & 14.0 & 6.5 & 17.8 & 26.0 & 15.7 & 11.3 & 16.4 & 23.9 \\
\tgroup{17}{Trained on the \bench{} graph (same triples, full-gallery InfoNCE)} \\
CLIP4Cir Combiner~\cite{baldrati2022clip4cir} (CLIP-L/14) & 15.3 & 9.9 & 16.3 & 25.1 & 7.8 & 4.4 & 7.3 & 13.3 & 28.0 & 21.1 & 31.0 & 40.5 & 20.4 & 14.8 & 22.0 & 30.2 \\
Qwen3-VL-Embedding-8B, LoRA fine-tuned & 21.3 & 15.3 & 21.9 & 33.0 & 12.9 & 8.6 & 12.7 & 20.4 & 35.5 & 28.1 & 38.2 & 49.1 & 27.2 & 20.7 & 28.9 & 39.6 \\
Fine-tuned emb. + name verifier (untuned Qwen3.5-27B) & 25.9 & 20.4 & 27.5 & 35.6 & 14.3 & 10.3 & 14.6 & 22.1 & 38.3 & 31.5 & 41.3 & 49.9 & 29.2 & 23.1 & 31.2 & 40.5 \\
Fine-tuned emb. + name verifier (\method{}) & 28.1 & 23.3 & 29.5 & 36.0 & 13.5 & 9.5 & 13.1 & 20.3 & 39.9 & 33.7 & 42.5 & 50.9 & 28.9 & 23.4 & 29.8 & 40.1 \\
Fine-tuned emb. + listwise verifier (\method{}) & 27.4 & 22.3 & 28.7 & 36.3 & \underline{15.7} & \underline{11.7} & \textbf{16.4} & \underline{22.4} & 41.4 & 35.6 & 43.9 & 51.6 & \textbf{30.9} & \textbf{25.0} & \underline{32.7} & \underline{41.7} \\
\tgroup{17}{Generative MLLM (Qwen3.5-27B) + name matching} \\
Generate answer $\to$ match & 6.8 & 5.4 & 6.7 & 9.2 & 5.4 & 3.9 & 5.5 & 8.2 & 14.9 & 12.3 & 15.2 & 19.5 & 12.1 & 10.0 & 12.2 & 15.2 \\
Generate with graph context (our linker) & 16.7 & 15.1 & 17.2 & 19.1 & 6.0 & 4.5 & 6.0 & 8.6 & 23.5 & 20.3 & 24.5 & 28.8 & 12.0 & 9.7 & 12.4 & 16.0 \\
\tgroup{17}{Knowledge-graph embedding} \\
ComplEx-N3~\cite{lacroix2018n3}, re-ID link & 23.4 & 20.1 & 24.5 & 29.2 & 5.9 & 4.0 & 5.9 & 8.8 & 32.1 & 25.8 & 34.5 & 44.2 & 17.4 & 13.2 & 18.6 & 24.8 \\
\textcolor{gray}{ComplEx-N3, oracle entity ID$^\dagger$} & 36.2 & 32.8 & 38.0 & 41.3 & 0.0 & 0.0 & 0.0 & 0.0 & 37.5 & 31.0 & 40.5 & 49.4 & 1.1 & 0.0 & 0.4 & 5.2 \\
\textcolor{gray}{NativE~\cite{zhang2024native}, oracle entity ID$^\dagger$} & 36.9 & 32.9 & 38.9 & 43.1 & 4.0 & 1.8 & 3.6 & 7.5 & 37.8 & 30.0 & 42.0 & 52.1 & 13.5 & 9.4 & 14.0 & 20.7 \\
\tgroup{17}{\method{} (ours, Qwen3.5-27B)} \\
Stage 1: frozen answer-onset states & 21.4 & 15.5 & 23.1 & 32.6 & 9.0 & 5.5 & 8.8 & 15.8 & 31.9 & 24.8 & 34.9 & 44.0 & 20.5 & 15.8 & 21.4 & 29.4 \\
Recall only & 31.3 & 26.0 & 33.2 & 41.3 & 9.8 & 6.1 & 9.7 & 16.8 & 37.7 & 30.8 & 40.0 & 50.7 & 21.9 & 17.0 & 22.3 & 30.6 \\
Recall + name verification & 34.4 & 29.5 & 36.3 & 43.8 & 11.9 & 8.2 & 12.1 & 18.5 & 40.8 & 34.0 & 44.0 & 53.6 & 24.2 & 18.7 & 25.2 & 35.4 \\
Recall + name + listwise verification & 36.2 & \underline{31.4} & 38.0 & 44.8 & 14.2 & 10.8 & 14.7 & 20.0 & \underline{44.1} & \underline{37.9} & \underline{47.4} & 54.8 & 27.1 & 21.9 & 28.3 & 36.7 \\
Recall + structural prior + name verification & 35.0 & 29.7 & 37.1 & 44.6 & 12.3 & 8.5 & 12.5 & 19.7 & 41.9 & 34.9 & 45.1 & 55.2 & 24.6 & 18.8 & 26.0 & 35.4 \\
Recall + priors + name verification & \underline{36.9} & 31.2 & \underline{39.3} & \underline{47.3} & 14.5 & 10.4 & 14.2 & 21.8 & 43.3 & 36.1 & 45.9 & \underline{57.0} & 28.8 & 22.1 & 31.0 & \underline{41.7} \\
Recall + priors + name + listwise (full system) & \textbf{39.1} & \textbf{33.9} & \textbf{41.2} & \textbf{48.0} & \textbf{16.2} & \textbf{12.3} & \underline{16.2} & \textbf{23.4} & \textbf{46.7} & \textbf{40.3} & \textbf{49.4} & \textbf{57.9} & \underline{30.6} & \underline{24.0} & \textbf{32.8} & \textbf{42.9} \\
\bottomrule
\end{tabular}}
\caption{\textbf{Full results} on \bench{}-Y/W test sets (filtered, full gallery, \%; \cref{tab:main} reports MRR and H@1).
Embedders receive the query photo and the relation instruction; for galleries we report the stronger of photo and photo+name.
$^\dagger$Uses the ground-truth entity ID of the query photo, which is unavailable in \bench{} and untrained for unseen entities; shown for reference. Best in bold, second best underlined.}
\label{tab:main_full}
\end{table*}

%% file: tables/oven.tex
\begin{table*}[t]
\centering
\small
\setlength{\tabcolsep}{5pt}
\begin{tabular}{l cccc cccc}
\toprule
 & \multicolumn{8}{c}{\bench{}-OVEN (in the wild)} \\
\cmidrule(lr){2-9}
 & \multicolumn{4}{c}{Transductive} & \multicolumn{4}{c}{Inductive} \\
Method & MRR & H@1 & H@3 & H@10 & MRR & H@1 & H@3 & H@10 \\
\midrule
\tgroup{9}{Zero-shot embedders} \\
CLIP-L/14~\cite{radford2021clip} (image + relation text) & 4.1 & 1.9 & 4.0 & 8.0 & 4.4 & 2.3 & 4.6 & 8.6 \\
SigLIP2-so400m~\cite{tschannen2025siglip2} (image + text) & 4.4 & 1.9 & 5.0 & 8.7 & 4.5 & 2.4 & 4.5 & 7.8 \\
BGE-VL-L (MegaPairs)~\cite{zhou2025megapairs} & 1.8 & 0.4 & 0.7 & 4.7 & 2.9 & 1.4 & 2.7 & 4.7 \\
GME-Qwen2-VL-7B~\cite{zhang2025gme} & 24.5 & 16.8 & 27.0 & 40.0 & 19.4 & 14.2 & 20.5 & 29.8 \\
Qwen3-VL-Embedding-8B~\cite{li2026qwen3vlemb} & 23.5 & 14.6 & 27.6 & 40.9 & 20.3 & 13.2 & 22.1 & 33.7 \\
\tgroup{9}{Trained on the \bench{} graph (same triples, full-gallery InfoNCE)} \\
CLIP4Cir Combiner~\cite{baldrati2022clip4cir} (CLIP-L/14) & 34.6 & 25.6 & 37.5 & 52.7 & 25.2 & 17.4 & 28.0 & 39.4 \\
Qwen3-VL-Embedding-8B, LoRA fine-tuned & 42.3 & 31.9 & 47.0 & 63.2 & 32.5 & 23.4 & 36.3 & 50.6 \\
Fine-tuned emb. + name verifier (untuned Qwen3.5-27B) & 42.6 & 31.9 & 47.9 & 63.7 & 33.0 & 24.0 & 36.2 & 50.6 \\
Fine-tuned emb. + name verifier (\method{}) & 43.3 & 33.1 & 48.2 & 64.0 & 32.4 & 23.4 & 35.8 & 50.2 \\
Fine-tuned emb. + listwise verifier (\method{}) & 50.6 & 41.2 & 56.3 & 67.8 & \underline{37.2} & \underline{27.8} & \underline{42.5} & \textbf{54.9} \\
\tgroup{9}{Generative MLLM (Qwen3.5-27B) + name matching} \\
Generate answer $\to$ match & 20.2 & 13.9 & 22.9 & 30.9 & 17.4 & 13.0 & 18.3 & 25.2 \\
Generate with graph context (our linker) & 27.1 & 19.9 & 32.1 & 36.9 & 16.5 & 12.3 & 17.1 & 24.4 \\
\tgroup{9}{Knowledge-graph embedding} \\
ComplEx-N3~\cite{lacroix2018n3}, re-ID link & 25.3 & 18.9 & 28.4 & 37.1 & 16.3 & 12.1 & 17.3 & 24.1 \\
\textcolor{gray}{ComplEx-N3, oracle entity ID$^\dagger$} & 25.6 & 19.1 & 27.8 & 39.0 & 3.4 & 1.8 & 3.6 & 6.5 \\
\textcolor{gray}{NativE~\cite{zhang2024native}, oracle entity ID$^\dagger$} & 34.5 & 26.4 & 38.1 & 49.0 & 17.4 & 11.7 & 18.7 & 27.9 \\
\tgroup{9}{\method{} (ours, Qwen3.5-27B)} \\
Stage 1: frozen answer-onset states & 34.8 & 25.4 & 37.7 & 52.6 & 19.6 & 12.8 & 20.6 & 32.2 \\
Recall only & 44.1 & 35.5 & 46.5 & 61.7 & 20.7 & 14.1 & 21.8 & 34.3 \\
Recall + name verification & 45.0 & 36.8 & 47.6 & 60.9 & 22.9 & 15.9 & 25.0 & 36.2 \\
Recall + name + listwise verification & \underline{53.7} & \underline{45.6} & \underline{59.2} & 66.9 & 29.2 & 22.7 & 31.8 & 40.4 \\
Recall + structural prior + name verification & 43.2 & 34.0 & 47.6 & 60.0 & 23.6 & 16.9 & 25.7 & 36.3 \\
Recall + priors + name verification & 46.8 & 35.6 & 51.7 & \underline{69.1} & 31.0 & 21.8 & 33.5 & 49.5 \\
Recall + priors + name + listwise (full system) & \textbf{56.1} & \textbf{46.1} & \textbf{63.4} & \textbf{72.2} & \textbf{38.0} & \textbf{28.6} & \textbf{42.7} & \underline{54.2} \\
\bottomrule
\end{tabular}
\caption{\textbf{Full results on \bench{}-OVEN} test set (filtered, full gallery, \%; \cref{tab:main} reports MRR and H@1). Rows and settings as in \cref{tab:main_full}.
$^\dagger$Uses the ground-truth entity ID of the query photo, which is unavailable in \bench{} and untrained for unseen entities; shown for reference. Best in bold, second best underlined.}
\label{tab:oven_full}
\end{table*}

%% file: tables/seeds.tex
\begin{table*}[h]
\centering
\small
{%
\begin{tabular}{l cc cc cc}
\toprule
 & \multicolumn{2}{c}{\bench{}-Y} & \multicolumn{2}{c}{\bench{}-W} & \multicolumn{2}{c}{\bench{}-OVEN} \\
 & Known & Unseen & Known & Unseen & Known & Unseen \\
\midrule
Recall only & 31.5$\pm$0.1 & 9.7$\pm$0.0 & 37.6$\pm$0.1 & 22.0$\pm$0.4 & 44.1$\pm$0.3 & 20.7$\pm$0.1 \\
Recall + name verification (top-20) & 34.2$\pm$0.2 & 11.6$\pm$0.1 & 40.5$\pm$0.0 & 23.7$\pm$0.2 & 44.8$\pm$0.6 & 23.2$\pm$0.5 \\
\bottomrule
\end{tabular}}
\caption{Test MRR (\%) of the recall model (final stage-2 configuration) over 3 training seeds, with and without top-20 name verification (mean$\pm$std).}
\label{tab:seeds}
\end{table*}

%% file: tables/sweep.tex
\begin{table*}[h]
\centering
\small
\setlength{\tabcolsep}{3.4pt}
{%
\begin{tabular}{l cccc cccc}
\toprule
 & \multicolumn{8}{c}{\bench{}-W} \\
 & \multicolumn{4}{c}{Trans.} & \multicolumn{4}{c}{Ind.} \\
 & MRR & H@1 & H@3 & H@10 & MRR & H@1 & H@3 & H@10 \\
\midrule
default ($r{=}16$, $\lambda{=}0.5$, lr $10^{-4}$, 2 ep.) & 35.0 & 28.3 & 37.6 & 47.6 & 21.0 & 16.2 & 21.9 & 29.9 \\
\quad + name verification (top-10) & 37.0 & 31.0 & 39.4 & 47.6 & 22.0 & 17.1 & 23.1 & 29.9 \\
LoRA $r{=}32$ & 35.0 & 28.3 & 37.7 & 47.8 & 21.3 & 16.6 & 21.6 & 30.2 \\
\quad + name verification (top-10) & 37.2 & 31.3 & 39.4 & 47.8 & 22.3 & 17.7 & 23.1 & 30.2 \\
$\lambda{=}1$ & 35.7 & 29.0 & 38.4 & 48.0 & 21.7 & 16.9 & 22.5 & 30.5 \\
\quad + name verification (top-10) & 37.9 & 31.9 & 40.3 & 48.0 & 22.9 & 18.0 & 24.0 & 30.5 \\
lr $2{\times}10^{-4}$ & 35.4 & 28.9 & 37.6 & 47.7 & 21.6 & 16.7 & 22.4 & 30.2 \\
\quad + name verification (top-10) & 37.2 & 31.2 & 39.2 & 47.7 & 22.8 & 18.1 & 23.9 & 30.2 \\
3 epochs & 35.2 & 28.5 & 37.6 & 47.3 & 21.4 & 16.5 & 22.4 & 30.4 \\
\quad + name verification (top-10) & 37.4 & 31.5 & 39.7 & 47.3 & 22.5 & 17.8 & 23.8 & 30.4 \\
\bottomrule
\end{tabular}}
\caption{Stage-2 hyper-parameters on \bench{}-W (2 GPUs with gradient accumulation, global batch 64; test, \%).}
\label{tab:sweep}
\end{table*}

%% file: tables/verifyk.tex
\begin{table*}[h]
\centering
\small
\setlength{\tabcolsep}{3.4pt}
{%
\begin{tabular}{l cccc cccc cccc cccc}
\toprule
 & \multicolumn{8}{c}{\bench{}-Y} & \multicolumn{8}{c}{\bench{}-W} \\
 & \multicolumn{4}{c}{Trans.} & \multicolumn{4}{c}{Ind.} & \multicolumn{4}{c}{Trans.} & \multicolumn{4}{c}{Ind.} \\
 & MRR & H@1 & H@3 & H@10 & MRR & H@1 & H@3 & H@10 & MRR & H@1 & H@3 & H@10 & MRR & H@1 & H@3 & H@10 \\
\midrule
Recall only & 31.3 & 26.0 & 33.2 & 41.3 & 9.8 & 6.1 & 9.7 & 16.8 & 37.7 & 30.8 & 40.0 & 50.7 & 21.9 & 17.0 & 22.3 & 30.6 \\
top-5, mean & 32.4 & 27.8 & 33.7 & 41.3 & 10.7 & 7.3 & 10.7 & 16.8 & 39.4 & 33.4 & 41.3 & 50.7 & 22.0 & 17.1 & 22.9 & 30.6 \\
top-5, sum & 32.5 & 27.9 & 33.9 & 41.3 & 10.6 & 7.2 & 10.6 & 16.8 & 39.3 & 33.4 & 41.3 & 50.7 & 22.4 & 17.6 & 23.1 & 30.6 \\
top-10, mean & 33.2 & 28.4 & 34.9 & 41.3 & 11.2 & 7.8 & 11.7 & 16.8 & 39.7 & 33.6 & 41.8 & 50.7 & 22.9 & 18.1 & 24.1 & 30.6 \\
top-10, sum & 33.3 & 28.5 & 35.0 & 41.3 & 11.0 & 7.7 & 11.3 & 16.8 & 40.0 & 33.9 & 42.4 & 50.7 & 23.2 & 18.3 & 24.1 & 30.6 \\
top-20, mean & 33.9 & 29.1 & 35.4 & 42.5 & 11.6 & 7.9 & 12.3 & 18.0 & 40.2 & 34.0 & 42.4 & 51.5 & 23.6 & 18.7 & 24.5 & 32.3 \\
top-20, sum & 34.1 & 29.3 & 36.1 & 42.5 & 11.5 & 7.9 & 11.9 & 18.0 & 40.5 & 34.2 & 43.1 & 52.1 & 23.9 & 19.0 & 24.7 & 33.2 \\
top-50, mean & 34.6 & 29.6 & 36.4 & 43.7 & 11.8 & 8.0 & 12.6 & 19.0 & 40.5 & 34.1 & 42.9 & 51.6 & 24.1 & 18.5 & 25.7 & 34.9 \\
top-50, sum & 34.4 & 29.5 & 36.3 & 43.8 & 11.9 & 8.2 & 12.1 & 18.5 & 40.8 & 34.0 & 44.0 & 53.6 & 24.2 & 18.7 & 25.2 & 35.4 \\
\bottomrule
\end{tabular}}
\caption{Verification budget: top-$K$ candidates and token aggregation (final model, test, \%). Weights are chosen on validation.}
\label{tab:verifyk}
\end{table*}